\pdfoutput=1
\documentclass[11pt]{article}
\usepackage{enumitem}
\usepackage[final]{acl}
\usepackage{float}
\usepackage{soul}
\usepackage{graphicx} 
\usepackage{times}
\usepackage{latexsym}
\usepackage{amsmath}
\usepackage{longtable}
\usepackage[normalem]{ulem}
\usepackage{booktabs} 
\usepackage[T1]{fontenc}

\usepackage[utf8]{inputenc}
\usepackage{adjustbox}
\usepackage{microtype}
\usepackage{multirow}
\usepackage{easyReview}
\usepackage{tcolorbox}
\usepackage{inconsolata}
\usepackage{pifont} 
\usepackage{adjustbox} 
\usepackage{booktabs} 
\usepackage{amsmath} 
\usepackage{pifont} 
\usepackage{adjustbox} 
\usepackage{booktabs} 
\usepackage{xcolor} 
\usepackage{xcolor, colortbl}
\definecolor{customPurple}{RGB}{148, 0, 211}
\definecolor{customGreen}{RGB}{0, 196, 159}
\definecolor{BEE7E9}{RGB}{190,231,233}
\definecolor{ECAD9E}{RGB}{236,173,158}
\definecolor{softpink}{RGB}{255,217,218}
\usepackage{listings}
\usepackage{tabularx}
\usepackage{makecell}
\usepackage{multirow}
\usepackage{array}
\usepackage{caption}
\usepackage{pdfpages}
\usepackage{booktabs} 
\usepackage{CJKutf8}
\usepackage[absolute,overlay]{textpos}
\definecolor{darkgreen}{rgb}{0.0, 0.5, 0.0}

\definecolor{softblue}{RGB}{40, 176 , 244} 
\definecolor{softyellow}{RGB}{255, 255, 150} 
\definecolor{lightgreen}{RGB}{145, 204, 117}

\newcommand{\greyhighlight}[1]{\colorbox{gray!30}{#1}}

\newcommand{\cmark}{\textcolor{darkgreen}{\ding{51}}} 
\newcommand{\xmark}{\textcolor{red}{\ding{55}}} 
\usepackage{colortbl}  
\usepackage{xcolor}  

\newcommand{\hmark}{\textcolor{blue}{\ding{51}\rotatebox[origin=c]{-6.2}{\kern-0.7em\ding{55}}}}

\newcommand{\ours}{\textsc{CANDY}}

\title{CANDY: Benchmarking LLMs' Limitations and Assistive Potential in Chinese Misinformation Fact-Checking}

\author{
  Ruiling Guo$^{1\#}$ \quad Xinwei Yang$^{2\#}$ \quad Chen Huang$^{23\diamondsuit}$ \quad Tong Zhang$^{2}$ \quad Yong Hu$^{1}$\thanks{Corresponding authors} \\
  $^1$School of Cyber Science and Engineering, Sichuan University, China\\
  $^2$College of Computer Science, Sichuan University, China\\
  $^3$Institute of Data Science, National University of Singapore, Singapore\\
  \texttt{\{ruiling\_guo, xinwei\_yang\}@stu.scu.edu.cn} \quad \texttt{huang\_chen@nus.edu.sg} 
}

\begin{document}

\begin{CJK}{UTF8}{gbsn}
\maketitle
\def\thefootnote{$\#$}\footnotetext{Both authors contributed equally to this study.}\def\thefootnote{\arabic{footnote}}

\def\thefootnote{$\diamondsuit$}\footnotetext{Work done during his PhD program.}\def\thefootnote{\arabic{footnote}}
\begin{abstract}

The effectiveness of large language models (LLMs) to fact-check misinformation remains uncertain, despite their growing use. To this end, we present \texttt{CANDY}, a benchmark designed to systematically evaluate the capabilities and limitations of LLMs in fact-checking Chinese misinformation. Specifically, we curate a carefully annotated dataset of $\sim$20k instances. Our analysis shows that current LLMs exhibit limitations in generating accurate fact-checking conclusions, even when enhanced with chain-of-thought reasoning and few-shot prompting. To understand these limitations, we develop a taxonomy to categorize flawed LLM-generated explanations for their conclusions and identify factual fabrication as the most common failure mode. Although LLMs alone are unreliable for fact-checking, our findings indicate their considerable potential to augment human performance when deployed as assistive tools in scenarios. Our dataset and code can be accessed at \url{https://github.com/SCUNLP/CANDY}.

\end{abstract}

\section{Introduction}

\begin{table}[t]

\renewcommand{\arraystretch}{1.2}

\small
\centering
    \begin{tabular}{p{0.45\textwidth}}
    \toprule
    \textbf{\textsc{Claim:}} A criminal case occurred in Jilin Province, where a butcher killed 12 people due to a land dispute.
    \\ 
    \sethlcolor{BEE7E9}
    \textbf{\textsc{LLM Response:}} Not a rumor. A criminal case \hl{did occur} in Jilin Province, attracting significant local attention. 
    \sethlcolor{softpink}
    According to the \hl{police report}, the suspect developed resentment from a land dispute, leading to the tragedy. \hl{Local government} confirmed there were 12 victims... 
    \\ 
    \midrule
    \textbf{\textsc{Annotation}} \\
    \textbf{Domain}: Society \\
    \textbf{Publish Date}: 2024/8/14\\
    \textbf{Is Misinformation}: Yes\\
    \sethlcolor{BEE7E9}
    \textbf{Gold Evidence}: After investigation by the police from Jilin, it was found that \hl{no such case occurred} at the location, and this information was entirely fabricated by ...\\ 
    \textbf{Source}: China Internet Joint Rumor Debunking Platform \\ 
    \textbf{Error Type of \textsc{LLM Response}}: Factual Fabrication \\ 
    \bottomrule 
    \end{tabular}
    \setlength{\abovecaptionskip}{0pt}
    \setlength{\belowcaptionskip}{0pt}
    \sethlcolor{softpink}
    \caption{An example from the \texttt{CANDYSET}, showing LLMs' tendency to sycophantize misinformation and \hl{fabricate authoritative verification}.}
    \label{tab:first example}
    \vspace{-5mm}
\end{table}

Misinformation, defined as \textit{"false or misleading information masquerading as legitimate news, regardless of intent"}\cite{glockner2022missing}. With the largest Internet user base, China produces vast amounts of such misinformation daily, making manual verification both labor-intensive and increasingly infeasible \cite{ChinaDaily, piyao, nakov2021automated}. 

Large language models (LLMs) possess extensive knowledge and explanatory capabilities, positioning them as promising tools for misinformation detection \cite{guan2023language, kang2024LLM_fault_localization, patil2024review_llms}. However, concerns about their effectiveness persist, such as hallucination limitations \cite{wang2024explainable,hu2024bad,augenstein2024factuality,fang2025lastingbench}. This has spurred numerous researches aimed at evaluating LLMs' ability to identify misinformation \cite{hoes2023leveraging, hsu2024enhancing, cekinel2024explaining, kao2024how, vykopal2024generative}. However, current evaluations are often superficial, primarily assessing the correctness of final answers (i.e., whether a statement is factual) while neglecting a deeper analysis of the characteristics and reasoning processes underlying LLMs' decisions. Consequently, it remains unclear precisely which types of misinformation LLMs can effectively identify, and what challenges they persistently face in this domain. Therefore, a comprehensive understanding of LLMs' capabilities and limitations in fact-checking Chinese misinformation remains uncertain, raising concerns about their practical utility.

To this end, we introduce \texttt{CANDY}, a benchmark designed to systematically evaluate the capabilities, limitations, and practical roles of LLMs in fact-checking  Chinese misinformation. To achieve this, we construct \texttt{CANDYSET}, a curated multi-domain dataset consisting of approximately 20k news instances collected from mainstream Chinese rumor refutation websites, with detailed sources provided in Table \ref{tab:top10source}. The dataset is further partitioned based on the cut-off dates of different models to support contamination-free evaluation (i.e., evaluating performance on unseen data).

With \texttt{CANDYSET}, we further expand our benchmark analysis using three tasks including \textit{Fact-Checking Conclusion} (i.e., if a statement is factual), \textit{Fact-Checking Explanation} (if a LLM-generated explanation for its fact-checking conclusion is reliable), and \textit{LLM-Assisted Fact-Checking} (to what extent can LLMs assist humans in fact-checking). To better identify and expose the deficiency of LLMs, we develop a taxonomy that categorizes flawed LLM-generated explanations into three dimensions, which are further divided into seven fine-grained categories (cf. Section \ref{tax}). Using this taxonomy, we manually annotate around 5k LLM-generated explanations to analyze explanation deficiencies, contributing a valuable component of our dataset (cf. Table \ref{data_statistics}). Finally, to evaluate the practical role of LLMs at real-world scenarios, our benchmark includes a human study to explore how LLMs can support users in fact-checking tasks.


\begin{table}[!ht]
\large
\resizebox{\linewidth}{!}{
\centering
\begin{tabular}{lccc}
\toprule
\multicolumn{1}{l}{\textit{\textbf{Basic Dataset Information}}} & \textbf{Real} & \textbf{Fake} & \textbf{Total} \\
\midrule
\# \textbf{Time Period} & \multicolumn{3}{c}{Mar. 2017 \textasciitilde~Oct. 2024} \\
\textbf{Total. \#Entries}  &10497	&9938	&20435 \\
\textbf{Avg. \#Claim Length (Tokens)}    &27.8  &33.6 & 30.6\\
\textbf{Avg. \#Gold Evidence Length (Tokens)}   &61.7 &65.3  &63.5    \\
\midrule
\# \textbf{Domain} &  &  &  \\
\cline{1-1}
\textit{Knowledge-intensive:} Politics  &822 &531 &1353 \\
\textit{Knowledge-intensive:} Culture  &1246 &323 & 1569 \\
\textit{Knowledge-intensive:} Science  &371 & 508 & 879 \\
\textit{Knowledge-intensive:} Health  &2849 & 3940 & 6789 \\
\textit{Temporal-sensitive:}  Society  &4270  &3633 &7903 \\
\textit{Temporal-sensitive:} Disasters  &625 & 604 & 1229 \\
\textit{Commonsense-sensitive:} Life     &314 &399 & 713 \\
\midrule
\multicolumn{1}{l}{\textit{\textbf{Annotated Flawed Fact-checking Explanations}}}   & \textbf{Correct} & \textbf{Wrong} & \textbf{Total}   \\
\midrule
\textbf{Total. \#Annotations}         &428   &4463   &4891 \\
\textbf{Avg. \#LLM Explanations Length (Tokens)}    &225.9   &206.2   &207.9      \\
\midrule
\# \textbf{Explanations Categorized by Taxonomy} &  &  &    \\  
\cline{1-1}
\textit{Faithful Hallucination: }Instruction Inconsistency &0  &5  &5 \\
\textit{Faithful Hallucination:} Logical Inconsistency   &72   &242   &314      \\
\textit{Faithful Hallucination}: Context Inconsistency   &68   &294    &362     \\
\textit{Factuality Hallucination:} Factual Fabrication    &132   &1571  &1703     \\
\textit{Factuality Hallucination:} Factual Inconsistency    &71   &1269    &1340    \\
\textit{Reasoning Inadequacy:} Overgeneralized Reasoning &58    &658    & 716       \\
\textit{Reasoning Inadequacy:} Under Informativeness     &27   &424   &451     \\
\bottomrule
\end{tabular}}
\setlength{\abovecaptionskip}{1pt}
\setlength{\belowcaptionskip}{1pt}
\caption{{\small Dataset statistics. 
"Correct" ("Wrong") indicates LLM achieved correct (wrong) fact-checking result.}}
\label{data_statistics}
\vspace{-2mm}
\end{table}


Our benchmark on sixteen LLMs and three large reasoning models (LRMs) reveals 
several challenges
: (1) Despite employing techniques such as Chain-of-Thought reasoning and few-shot prompting, LLMs struggle with fact-checking conclusions, especially when addressing contamination-free evaluation and time-sensitive events. Our analysis on the fact-checking conclusion task showed how LRM accuracy and timeliness often fall short during societal crises or disasters, especially since there is no specific event feature in the dataset. (2) Furthermore, our analysis on the explanation generation task revealed the main reason why LLMs often produce incorrect fact-checking conclusions: their tendency toward factual hallucination (particularly true for LRMs). In some cases, they even fabricate highly deceptive details to support false claims, which significantly limits their ability to serve as independent Chinese fact-checking decision makers. (3) Finally, our LLM-assisted fact-checking task showed that, under identical testing conditions, individuals from diverse educational backgrounds achieve higher accuracy and efficiency when assisted by LLMs. This suggests that LLMs are best positioned to serve as intelligent assistants or advisors in the fact-checking process, rather than as autonomous decision-makers.

\begin{table*}[t]
\centering
\begin{adjustbox}{width=0.99\textwidth}
\begin{tabular}{l|c|cccc|cc|cc}
\toprule
\multirow{3}{*}{\textbf{\makecell[c]{Fact-checking Benchmark}}} & 
\multirow{3}{*}{\textbf{\makecell[c]{Language}}} &
\multicolumn{4}{c|}{\textbf{Basic Dataset Information}} & 
\multicolumn{2}{c|}{\textbf{Annotation for LLM Explanation}} & 
\multicolumn{2}{c}{\textbf{Human Study}} \\ \cline{3-10}
& & \textbf{\makecell{Claims}} & 
\textbf{\makecell{Time}} & 
\textbf{\makecell{Multi-\\Domain}} & 
\textbf{\makecell{Ground-truth\\ Evidence}} & 
\textbf{\makecell{Explanation \\ Annotations}} & 
\textbf{\makecell{Explanation \\ Taxonomy}} &
\textbf{\makecell{Various \\ Education Level}} &
\textbf{\makecell{Human-LLM \\ Interaction Records}} \\
\midrule

PolitiFact\cite{kao2024how}  & English &33721 &\xmark & \xmark & \xmark & \hmark &\xmark &\xmark & \xmark\\
FlawCheck\cite{hsu2024enhancing} & English &50 &2019-2021 &\xmark & \xmark &\hmark &\xmark & \xmark & \xmark\\
\midrule
Weibo\cite{jin2017multimodal} & Chinese & 9528 & May 2012–Jan 2016 & \cmark & \xmark & \xmark & \xmark & \xmark & \xmark \\
COVID19\cite{yang2022know} & Chinese & 623837 & Jan–May 2020 & \xmark & \hmark & \xmark & \xmark & \xmark & \xmark \\
CHECKED\cite{yang2020checked} & Chinese & 2104 & Dec 2019–Aug 2020 & \xmark & \xmark & \xmark & \xmark & \xmark & \xmark \\
CrossFake\cite{du2021cross} & Chinese & 219 & Feb–Dec 2020 & \xmark & \xmark & \xmark & \xmark & \xmark & \xmark \\
Weibo21\cite{nan2021mdfend} & Chinese & 9128 & Dec 2014–Mar 2021 & \cmark & \xmark & \xmark & \xmark & \xmark & \xmark \\
MCFEND\cite{li2024mcfend} & Chinese & 23974 & Mar 2015–Mar 2023 & \cmark & \xmark & \xmark & \xmark & \xmark & \xmark \\
LTCR\cite{ma2023ltcr} & Chinese & 2290 & Dec 2019–Jan 2023 & \xmark & \xmark & \xmark & \xmark & \xmark & \xmark \\
CHEF\cite{hu2022chef} &Chinese & 10000 & Sep 2017–Dec 2021 & \cmark & \cmark & \xmark & \xmark & \xmark & \xmark \\

\midrule
CANDYSET\textbf{(\textit{Ours})} &Chinese & 20435 & Mar 2017–Oct 2024 & \cmark & \cmark & \cmark & \cmark & \cmark & \cmark \\
\bottomrule
\end{tabular}
\end{adjustbox}
\setlength{\abovecaptionskip}{1pt}
\setlength{\belowcaptionskip}{1pt}
\caption{Advantages of \ours~over other benchmarks. Here, '\hmark'~indicates partial support.}
\vspace{-3mm}
\label{table:datasetcompare}
\end{table*}

In this paper, \texttt{CANDY} serves as a valuable resource for offering practical guidance and insights for LLM fact-checking. In conclusion, our contributions are as followings:
\begin{itemize}[leftmargin=*, itemindent=0.05cm, itemsep=-2pt]

\item We propose \texttt{CANDY}, the first benchmark provides a comprehensive evaluation and in-depth analysis of LLMs' ability to fact-check Chinese misinformation, as well as their applicability in practice.

\item We introduce \texttt{CANDYSET}, a Chinese large-scale dataset. It comprises \textasciitilde20k raw instances, \textasciitilde5k carefully annotated LLM-generated explanations, and \textasciitilde7k  human study samples. This enables in-depth evaluation.

\item We introduce a fine-grained taxonomy for categorizing flawed LLM explanations, facilitating in-depth analysis of LLM deficiencies in fact-checking misinformation.

\item With \texttt{CANDY}, we experimentally benchmark sixteen LLMs (and three LRMs) using three progressive tasks to systematically evaluate the capabilities, limitations, and practical roles of LLMs in fact-checking real-world Chinese misinformation, offering practical guidance for future study.

\end{itemize}

\section{Related Works}
In recent years, with the proliferation of misinformation on social media, several works have developed various fact-checking benchmarks as shown in Table \ref{table:datasetcompare}. For instance, COVID19-Health-Rumor \cite{yang2022know}, CHECKED \cite{yang2020checked} focus on rumors during public health crises, and Weibo21 \cite{nan2021mdfend} and MCFEND \cite{li2024mcfend} for multi-domain scenarios. However, except for CHEF\cite{hu2022chef}, these benchmarks lack detailed explanations of supporting evidence, limiting their usefulness for deeper analysis.  Additionally, of all the benchmarks, only PolitiFact\cite{kao2024how} and FlawCheck\cite{hsu2024enhancing} strive to get LLMs to generate explanations during the fact-checking process. However, both of these benchmarks fall short in providing detailed human annotations and thorough analysis of the accuracy or shortcomings of these explanations. Moreover, none of the benchmarks consider testing LLMs in collaboration with real users in practical scenarios, which hinders their real-world applicability. Our work addresses these limitations by introducing (1) carefully annotated evidence explanations for each instance, (2) constructing a taxonomy to categorize errors in LLM-generated explanations, and (3) introducing a real-world human study to examine the practical deployment of LLMs in authentic fact-checking scenarios.

\section{CANDY Benchmark}


\subsection{CANDYSET Dataset}
\label{Datades}
\textbf{Overview}. To facilitate in-depth Chinese fact-checking evaluation, We introduce \texttt{CANDYSET}, a large-scale dataset comprising three key components: approximately 20,000 multi-domain instances of both misinformation and authentic news, complete with fact-checking evidence; 4,891 manually annotated flawed LLM-generated fact-checking explanations; and records of human-LLM interactions during fact-checking. Detailed statistics of the dataset are shown in Table \ref{data_statistics}.

\noindent\textbf{Multi-domain Claims \& Evidences Collection}. Collected from authoritative Chinese fact-checking platforms (e.g., the China Internet United Rumor Refutation Platform\footnote{\href{https://www.piyao.org.cn}{https://www.piyao.org.cn}}, with additional sources listed in Table \ref{tab:top10source}) via HTML scrapers\footnote{Scraper code will be released along with our dataset.}, our dataset includes claims (i.e., news that may be misinformation), supporting evidence, publication dates, and domains. After data preprocessing, we manually link each claim with its gold evidence. Uniquely designed for contamination-free evaluation (i.e., evaluating performance on unseen data), the dataset spans March 2017 to October 2024 and is split by date according to each model's cut-off, thus enabling contamination-free performance simulation. Further details are provided in Appendix \ref{sec:dataconstruct}.

\noindent\textbf{Fact-checking Explanation Annotation}. To facilitate in-depth analysis of LLM fact-checking, we first required each LLM to generate explanations for its fact-checking decisions. We then manually compared these explanations with the collected evidence within our dataset to identify flaws. Due to the human cost involved, we considered explanations from eleven LLMs, each responsible for 2,000 randomly sampled entries from the raw dataset based on the original time and domain distribution. This resulted in 22,000 explanations, of which 4,891 were identified as flawed. Among these, only 428 led to correct LLM fact-checking outcomes, while the vast majority (4,463) resulted in incorrect outcomes. To provide more insights, we manually categorized these flawed explanations according to our pre-defined taxonomy (cf. Section \ref{tax}). During the whole process, ten computer science master's students, fluent in Chinese and experienced in data annotation, were involved. Each explanation was independently labeled by two annotators, and discrepancies were resolved by a third annotator. A Fleiss' Kappa of 0.76 indicates substantial agreement among the annotators, confirming the reliability of the annotation process. More details are provided in Appendix \ref{sec:dataconstruct}.


\noindent\textbf{Quality Control}. \texttt{CANDYSET}'s quality is rigorously controlled through two vetting layers: authoritative data sources, and human annotation and review process: (1) reliance on authoritative data sources, ensuring the credibility of news and evidence, and (2) a human annotation and review process, where LLM-generated fact-checking explanations are meticulously annotated and checked by multiple humans, resulting in substantial annotation agreement (Fleiss' Kappa of 0.76).

\subsection{Taxonomy for Fact-checking Explanations}
\label{tax}
We propose a fine-grained taxonomy to group the deficiencies of LLM generated explanations across various expression tones, ranging from confident and definitive to speculative. It encompasses the following three dimensions. Table \ref{tab:taxonomy} summarizes the taxonomy and provides examples.


\noindent\textbf{Faithfulness Hallucination} occurs when the LLM's explanation is unfaithful to the user input, or contains logical inconsistencies, questioning its meaningfulness. Inspired by \citet{huang2023survey}, we consider three subcategories.
\begin{itemize}[leftmargin=*,itemindent=0.05cm, itemsep=-2pt]
  \item \textit{Instruction Inconsistency}. The LLM's output deviates from the user's instructions, particularly when unrelated to fact-checking.
    \item \textit{Logical Inconsistency}. The LLM's output contains internal logical conflicts. For example, \textit{"Cristiano Ronaldo played for Real Madrid from 2009 to 2018, for a total of 8 years."}
     \item \textit{Context Inconsistency}. The LLM's output contradicts the user-provided context. For instance, the LLM misjudges the claim: "\textit{In the case of bacterial infection, antibiotics can be used to treat patients with COVID-19.}" as misinformation because it focuses solely on the latter part.
 \end{itemize}

\noindent\textbf{Factuality Hallucination} refers to the LLM expressing reasons in a definitive tone that contradict real-world facts or are fabricated \cite{huang2023survey, qin-etal-2024-beyond, deng2024towards}. We identify two subcategories:
\begin{itemize}[leftmargin=*,itemindent=0.05cm, itemsep=-2pt]
    \item \textit{Factual Fabrication} refers to the LLM's output that fabricates rationales for analysis without relying on any real-world information. For instance, it propagates misinformation by stating, "\textit{It was reported by reputable media outlets.}"
    \item \textit{Factual Inconsistency} refers to the LLM’s output contains facts that can be grounded in real-world information, but present contradictions. For example: \textit{"China will host the World Cup in 2026."}
\end{itemize}

\noindent\textbf{Reasoning Inadequacy} refers to the inability of an LLM to deliver high-quality and helpful reasoning when direct evidence is insufficient.
\begin{itemize}[leftmargin=*, itemindent=0.05cm, itemsep=-2pt]
    \item \textit{Overgeneralized Reasoning} refers to the tendency of a LLM to produce speculative rationales based on overly broad or superficial criteria. For example: Solely based on "\textit{the technology sector has indeed seen rapid advancements in recent years}," concluding that "\textit{the new technology can increase battery life by ten times.}"
    \item \textit{Under Informativeness} refers to the tendency of a LLM to exhibit excessive rigor or restraint, failing to provide more contextually valuable content. For example:\textit{ "There is currently no conclusive scientific evidence proving that eating an apple a day is beneficial to health."}
\end{itemize}

\section{Benchmark Setup}

\textbf{Overview}. We benchmark LLMs (and LRMs) using the following three tasks to systematically evaluate the capabilities, limitations, and practical roles of LLMs in fact-checking real-world Chinese misinformation: Fact-Checking Conclusion, Fact-Checing Explanation, and LLM-Assisted Fact-Checking, each detailed in the subsequent sections.




\noindent\textbf{Baselines}. Our study benchmarks sixteen LLMs, such as models from the OpenAI GPT family \cite{openai2023gpt} and the Qwen family \cite{qwen2.5}. Also, we consider three LRMs, including OpenAI O1-Mini \cite{jaech2024openai}, DeepSeek-R1 \cite{guo2025deepseek} and Qwen-QwQ \cite{qwq32b}. Following \citep{deng-etal-2023-prompting}, we devise different prompting schemes for each baseline: 1) Zero-shot w/o CoT. 2) Zero-shot w/ CoT \citep{wei2022chain}. 3) Few-shot w/o CoT \citep{dong2022survey}. 4) Few-shot w/ CoT\cite{Dong2022}. For detailed information about the models and prompt scheme design, please refer to Appendix \ref{sec:model_implementation} and Appendix \ref{sec:prompt}, respectively.

\noindent\textbf{Evaluation Metrics \& Implementation Details}. Specific metrics and implementation details for each task will be presented in corresponding sections. Further details are in Appendix \ref{detailsimp}. 

\section{Task 1: Fact-Checking Conclusion}
\label{sec:conclu}

This section aims to evaluate the ability of LLMs to verify facts by assessing their performance in distinguishing factual statements from falsehoods. Following \citet{huang2024from}, we adopt accuracy (Acc.) and F1 score as evaluation metrics.

\subsection{Overall Evaluation}
\label{task1over}

\begin{table*}[!ht]
\scriptsize
\centering
\resizebox{\textwidth}{!}{
\begin{tabular}{@{}l|cccccccccc@{}}
\toprule
\multirow{2}{*}{\textbf{Model} (Cut-off Date)}
& \multicolumn{2}{c|}{\textbf{Zero-shot w/o CoT}} & \multicolumn{2}{c|}{\textbf{Zero-shot w/ CoT}} & \multicolumn{2}{c|}{\textbf{Few-shot w/o CoT}} & \multicolumn{2}{c|}{\textbf{Few-shot w/ CoT}} & \multicolumn{2}{c}{\textbf{Average Performance}} \\ \cmidrule(lr){2-3} \cmidrule(lr){4-5} \cmidrule(lr){6-7} \cmidrule(lr){8-9} \cmidrule(lr){10-11}
& Acc. & F1 & Acc. & F1 & Acc. & F1 & Acc. & F1 & Acc. & F1 \\ 
\midrule
\multicolumn{11}{c}{\textit{\textbf{Closed Source Models}}} \\ \midrule

\textbf{GPT-4o (2023.11)}  & 74.3(\underline{\textbf{85.3}})	& \underline{\textbf{73.6}}(\underline{\textbf{83.7}})	& \underline{\textbf{76.8}}(\underline{\textbf{86.6}})	& \underline{\textbf{77.2}}(85.2) &	74.9(\underline{\textbf{85.7}})	& 75.1(\underline{\textbf{86.5}})	& \underline{\textbf{78.6}}(\underline{\textbf{87.1}})	& \underline{\textbf{78.8}}(\underline{\textbf{88.0}})	& \underline{\textbf{76.2}}(\underline{\textbf{86.2}}) &	\underline{\textbf{76.1}}(\underline{\textbf{85.9}})
 \\

\textbf{GPT-4-Turbo (2023.5)} & 72.2(81.1)	& 72.0(80.6) &	75.4(85.2)	& 73.5(83.2)&	73.1(82.3)&	71.8(84.8)&	75.5(85.1)	&74.2(84.0)	& 74.1(83.4) &	72.9(83.2)
   \\

\textbf{GPT-3.5-Turbo (2021.10)}  &65.2(78.3)	 &60.4(75.2)	&68.5(79.2)	 &64.3(76.6)	&66.6(81.7)	&59.3(86.3)	&70.2(82.2)	 &71.4(82.7)	&67.6(80.4)	 &63.9(80.2)
   \\

\textbf{Gemini-1.5-pro (2023.11)}  &74.3(79.3)  &72.1(77.1)	&73.5(78.3)	 &71.8(76.2) &	74.5(79.3)	&74.2(80.0)	 &76.7(83.5)	&77.3(84.0)	 &74.8(80.1)	&73.9(79.3)
  \\

\textbf{Baichuan4-Turbo (2024.4)}   &68.3(72.3)	&47.4(62.2)	 &69.5(74.4)	&48.5(63.1)	&66.4(70.5)	 &44.9(60.5)	&70.4(75.3)	 &53.1(65.2)	&68.7(73.1)	&48.5(62.8)
   \\

\textbf{Yi-large (2023.6)}  &70.2(72.2)	&68.5(71.5)	 &74.1(75.6)	&71.4(73.5)	&73.4(76.4)	&69.2(77.2)	&74.2(78.1)	&68.8(73.2)	&73.0(75.6)	&69.5(73.9) \\

\textbf{ChatGLM4 (2022.10)}   &\underline{\textbf{74.4}}(82.4)	 &70.3(81.1)	&76.0(85.2)	&72.3(\underline{\textbf{86.4}})	&\underline{\textbf{76.7}}(84.0)	&73.0(85.9)	&77.3(86.6)	&74.3(85.2)	&76.1(84.6)	&72.5(84.7) \\

\textbf{DeepSeek-v3 (2024.7)}  &72.2(79.8)	&73.1(81.2)	 &76.4(84.3)	&75.3(83.5)	&74.5(81.2)	 &\underline{\textbf{76.3}}(83.1)	&76.1(86.2)	 &76.5(85.5)	&74.8(82.9)	&75.3(83.3)
  \\

\midrule

\textbf{O1-Mini (2023.12)} &-- &-- &71.2(78.9) &70.6(80.2)  &-- &-- &72.0(80.4) &71.3(81.1)  &71.6(79.7)  &71.0(80.7)    \\
\textbf{DeepSeek-R1 (2024.7)} &-- &-- &73.4(82.3) &72.3(81.9)  &-- &-- &73.6(83.2) &74.2(84.9)  & 73.5(82.8) &73.3(83.4) \\
\textbf{Qwen-QwQ-Plus (2024.8)}  &-- &-- &73.8(81.6) &71.3(78.9)  &-- &-- &74.1(82.8) &73.6(83.1) &74.0(82.2) &72.5(81.0)\\

\midrule

\textbf{\textit{Average}} &71.4(78.8)	&67.1(76.6)	&73.5(81.1)	&69.9(79.0)	&72.5(80.1)	&68.0(80.5)	&74.4(82.8)	&72.1(81.5)	&73.1(81.9)	&69.9(79.9)  \\ 

\midrule

\multicolumn{11}{c}{\textit{\textbf{Open Source Models}}} \\ \midrule

\textbf{Yi-1.5-6B(2024.5)}   &60.5(63.3)	&35.8(36.8)	 &63.2(61.1)	&40.1(46.2)	&58.6(62.1)	&32.6(34.1)	&59.7(63.3)	&34.7(38.2)	&60.5(62.5)	&35.8(38.8)   \\

\textbf{Qwen-2.5-7B(2023.10)}  &62.3(64.2)	&29.4(30.4)	 &64.2(60.4)	&26.3(29.2)	&61.7(57.3)	&23.1(31.2)	&62.8(63.7)	&30.3(32.2)	&62.8(61.4)	&27.3(30.8) \\

\textbf{Llama-3.2-7B(2023.12)}   &58.6(62.2)	&30.3(34.2)	&57.2(61.1)	&30.1(33.4)	&57.1(60.5)	&29.2(32.9)	&60.3(65.2)	&32.4(36.8)	&58.3(62.3)	&30.5(34.3)  \\

\textbf{GLM4-9B(2023.10)}  &63.2(70.4)	&47.3(49.3)	 &68.3(74.1)	&49.2(48.2)	&67.2(72.4)	&45.6(41.4)	&70.2(74.6)	&52.3(56.3)	&67.2(72.9)	&48.6(48.8)  \\

\textbf{Yi-1.5-9B(2024.5)} &62.0(67.2)	&46.4(50.1)	&67.2(72.6)	&50.1(53.7)	&65.9(70.4)	&55.5(60.1)	&68.2(74.1)	&60.5(64.1)	&65.8(71.1)	&53.1(57) \\

\textbf{Qwen-2.5-14B(2023.10)} &68.2(73.1)	&69.1(71.5)	&67.1(71.2)	&68.0(71.1)	&71.1(75.6)	&67.8(72.5)	&74.2(78.1)	&71.4(77.3)	&70.2(74.5)	&69.1(73.1) \\

\textbf{Llama-3.2-70B(2023.12)} &70.3(78.6)	&\underline{\textbf{72.7}}(79.2)	&75.2(81.0)	&\underline{\textbf{73.8}}(81.7)	&73.1(79.6)	&70.5(77.8)	&76.2(82.5)	&71.3(79.8)	&73.7(80.4)	&72.1(79.6)  \\

\textbf{Qwen-2.5-72B(2023.10)} &\underline{\textbf{73.5}}(\underline{\textbf{80.1}})	&71.7(\underline{\textbf{80.3}})	&\underline{\textbf{75.3}}(\underline{\textbf{82.1}})	&73.4(\underline{\textbf{83.2}})	&\underline{\textbf{76.0}}(\underline{\textbf{83.2}})	&\underline{\textbf{72.6}}(\underline{\textbf{82.3}})	&\underline{\textbf{76.6}}(\underline{\textbf{85.6}})	&\underline{\textbf{77.8}}(\underline{\textbf{84.3}})	&\underline{\textbf{75.4}}(\underline{\textbf{82.8}})	&\underline{\textbf{73.9}}(\underline{\textbf{82.5}}) \\

\midrule

\textbf{\textit{Average}}  &64.8(69.9)	&50.3(54)	&67.2(70.5)	&51.4(55.8)	&66.3(70.1)	&49.6(54)	&68.5(73.4)	&53.8(58.6)	&66.7(71.0)	&51.3(55.6)\\

\midrule

\textit{\textbf{Average over all LLMs}} &68.1(74.4) 	&58.7(65.3) 	&70.9(76.6) 	&62.1(69.2) 	&69.4(75.1) 	&58.8(67.3) 	&71.9(78.8) 	&64.4(71.9) 	&70.4(77.3) 	&62.1(69.6) \\

\bottomrule
\end{tabular}} 
\setlength{\abovecaptionskip}{0pt}   
\setlength{\belowcaptionskip}{0pt}
\caption{Fact-checking conclusion performance (\%) on \texttt{CANDYSET}. Values outside the parentheses indicate performance on contamination-free evaluation, while values inside indicate performance on contamination evaluation.}
\label{tab:overall performance}
\vspace{-3mm}
\end{table*}

\begin{table}[ht]
\resizebox{\linewidth}{!}{
\begin{tabular}{ccc|cc|cc}
\toprule
\multirow{2}{*}{\makecell{\textbf{Model Type} }} & \multirow{2}{*}{\makecell{\textbf{Model}}} & \textbf{Zero-shot} & \textbf{Zero-shot} & \textbf{Difference} & \textbf{Few-shot} & \textbf{Difference} \\
& & \textbf{w/o CoT} & \textbf{w/ CoT} & \textbf{(COT)} & \textbf{w/o CoT} & \textbf{(Few-shot)} \\ \hline
\multirow{11}{*}{\makecell{\textbf{Closed} \\ \textbf{Source} \\ \textbf{Models}}}  
& GPT-4o        & 7.21  & 5.32  & -1.89 & 7.89  & +0.68 \\
& GPT-4-Turbo   & 10.68 & 14.46 & +3.78 & 12.57 & +1.89 \\
& GPT-3.5-Turbo & 12.12 & 8.31  & -3.81 & 16.86 & +4.74 \\
& Gemini-1.5-pro & 6.18 & 10.14 & +3.96 & 6.49  & +0.31 \\
& Baichuan4-Turbo & 15.35 & 24.67 & \greyhighlight{+9.32} & 20.27 & +4.92 \\
& Yi-large      & 12.33 & 16.47 & +4.14 & 18.72 & +6.39 \\
& ChatGLM4      & 7.49  & 6.77  & -0.72 & 10.22 & +2.73 \\
& DeepSeek-v3   & 8.92  & 13.23 & +4.31 & 7.84  & -1.08 \\
\hline
\multirow{8}{*}{\makecell{\textbf{Open} \\ \textbf{Source} \\ \textbf{Models}}}  
& Yi-1.5-6B     & 15.31 & 22.33 & \greyhighlight{+7.02} & 22.24 & \greyhighlight{+6.93} \\
& Qwen-2.5-7B   & 11.44 & 18.78 & \greyhighlight{+7.34} & 13.57 & +2.13 \\
& Llama-3.2-7B  & 14.38 & 18.29 & +3.91 & 22.58 & \greyhighlight{+8.20} \\
& GLM4-9B       & 11.75 & 24.33 & \greyhighlight{+12.58} & 19.32 & \greyhighlight{+7.57} \\
& Yi-1.5-9B     & 16.29 & 24.75 & \greyhighlight{+8.76} & 19.28 & +2.99 \\
& Qwen-2.5-14B  & 13.57 & 15.68 & +2.11 & 13.74 & +0.17 \\
& Llama-3.2-70B & 27.13 & 24.63 & -2.50 & 24.57 & +2.44 \\
& Qwen-2.5-72B  & 22.82 & 23.93 & +1.11 & 17.11 & -0.71 \\
\bottomrule
\end{tabular}}
\setlength{\abovecaptionskip}{0pt}   
\setlength{\belowcaptionskip}{0pt}
\caption{Overconfidence evaluation on LLMs with/without CoT and few-shot prompting. Significant differences are marked in \greyhighlight{grey}.}
\label{uncertainty_combined}
\vspace{-3mm}
\end{table}

As shown in Table \ref{tab:overall performance}, the GPT-4o emerged as the top-performing model, which may underscores its robust utilization of extensive internal knowledge. 
Our detailed observations are as follows:

\begin{figure}[t]
	\centering
	\includegraphics[width=0.47\textwidth]{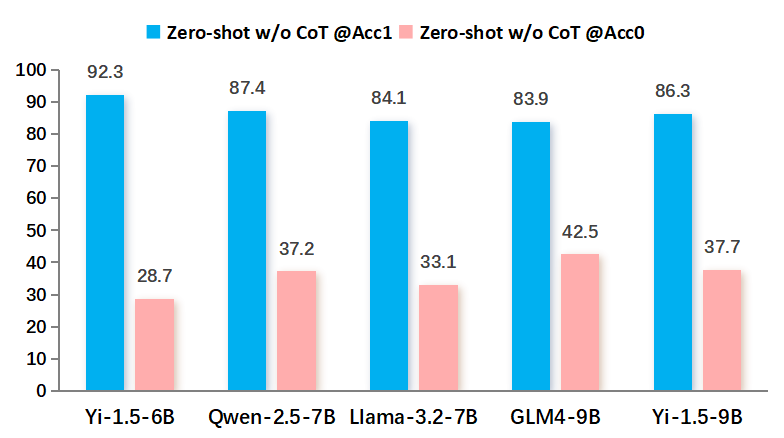}
    \setlength{\abovecaptionskip}{1pt}
     \setlength{\belowcaptionskip}{1pt}
	\caption{Fact-checking accuracy when handling authentic claims (Acc@0) and misinformation (Acc@1). LLMs tend to classify data as misinformation. }
	\label{fig:pro_con}
    \vspace{-5mm}
\end{figure}

\noindent\textbf{Current LLMs, even when employing methods like CoT reasoning and few-shot prompting, still struggle to accurately perform fact-checking tasks, particularly in contamination-free scenarios.} While larger models such as Llama-3.2-70B, Qwen-2.5-72B, and GPT-4o demonstrate higher performance, even the top-performing GPT-4o only achieves moderate results (76.2\% accuracy and 76.1\% F1 score). Notably, LLM performance declines significantly when handling contamination-free evaluation compared to contamination evaluation, with an average decrease of 6.9\% in accuracy and 7.5\% in F1 score. This performance gap highlights the complexities of contamination-free fact-checking, which requires dynamic assessment of rapidly evolving information, unlike contamination fact-checking that often relies on static, pre-verified data. Smaller-scale open-source LLMs (e.g., Yi-1.5-6B, Llama-3.2-7B, Qwen-2.5-7B) exhibit even lower performance, often misclassifying truthful information as misinformation, leading to a considerable gap between accuracy and F1 (e.g., Yi-1.5-6B: 60.5\% accuracy, 35.8\% F1 in Table \ref{tab:overall performance}). The reason for this gap is illustrated in Figure \ref{fig:pro_con}, small-scale LLMs tend to classify most data instances as misinformation. Furthermore, methods like CoT reasoning and few-shot prompting may exacerbate overconfidence issues in small-scale open-source models, leading to adverse outcomes. Indeed, our analysis in Table \ref{uncertainty_combined} using Expected Calibration Error \citet{cole-etal-2023-selectively} reveals that CoT and few-shot prompting often lead overconfidence, while simultaneously less accurate in detecting misinformation, thereby counteracting the intended improvements.

\begin{table*}[ht]
    \centering
    \scriptsize
    \begin{tabularx}{\textwidth}{l@{\hspace{8pt}}*{14}{>{\centering\arraybackslash}X@{\hspace{8pt}}}}
    \toprule
    \multirow{3}{*}{\textbf{Methods}} 
    & \multicolumn{4}{@{}c@{\hspace{10pt}}}{\textbf{Temporal-sensitive}} 
    & \multicolumn{8}{@{}c@{\hspace{10pt}}}{\textbf{Knowledge-intensive}} 
    & \multicolumn{2}{@{}c@{\hspace{10pt}}}{\textbf{Commonsense}} \\
    \cmidrule(l{4pt}r{4pt}){2-5} \cmidrule(l{4pt}r{4pt}){6-13} \cmidrule(l{4pt}r{4pt}){14-15}
    & \multicolumn{2}{@{}c@{}}{\textbf{Society}} 
    & \multicolumn{2}{@{}c@{}}{\textbf{Disasters}} 
    & \multicolumn{2}{@{}c@{}}{\textbf{Health}} 
    & \multicolumn{2}{@{}c@{}}{\textbf{Politics}} 
    & \multicolumn{2}{@{}c@{}}{\textbf{Culture}} 
    & \multicolumn{2}{@{}c@{}}{\textbf{Science}} 
    & \multicolumn{2}{@{}c@{}}{\textbf{Life}} \\
    \cmidrule{2-3} \cmidrule{4-5} \cmidrule{6-7} \cmidrule{8-9} \cmidrule{10-11} \cmidrule{12-13} \cmidrule{14-15}
    & {Acc.} & {F1} & {Acc.} & {F1} & {Acc.} & {F1} & {Acc.} & {F1} & {Acc.} & {F1} & {Acc.} & {F1} & {Acc.} & {F1} \\
    \midrule
    \textbf{GPT-4o} & 76.92 & \underline{\textbf{78.21}} & 74.21 & \underline{\textbf{73.78}} & \underline{\textbf{87.97}} & \underline{\textbf{89.66}} & 84.85 & 82.64 & 82.98 & 67.08 & \underline{\textbf{82.59}} & \underline{\textbf{85.41}} & \underline{\textbf{82.19}} & \underline{\textbf{84.61}} \\
    \textbf{GPT-4-Turbo} & 72.57 & 74.14 & 72.66 & 72.95 & 83.80 & 86.15 & 75.54 & 73.96 & 77.50 & 57.42 & 81.46 & 84.10 & 74.75 & 77.56 \\
    \textbf{GPT-3.5-Turbo} & 67.83 & 66.99 & 64.77 & 63.02 & 75.36 & 77.30 & 71.62 & 69.33 & 74.63 & 51.23 & 74.86 & 76.11 & 62.13 & 58.20 \\
    \textbf{Baichuan4-Turbo} & 70.69 & 58.07 & 60.13 & 37.34 & 66.58 & 60.72 & 79.67 & 71.38 & 83.62 & 44.73 & 67.35 & 61.58 & 57.36 & 38.96 \\
    \textbf{ChatGLM4} & 79.08 & 76.60 & \underline{\textbf{74.34}} & 71.93 & 83.35 & 84.30 & 83.73 & 79.92 & 86.95 & 66.67 & 80.75 & 81.77 & 70.99 & 68.79 \\
    \textbf{DeepSeek-V3} & 78.82 & 78.12 & 73.16 & 72.82 & 86.78 & 88.22 & 87.13 & \underline{\textbf{86.64}} & 88.04 & \underline{\textbf{76.71}} & 82.54 & 84.12 & 82.02 & 82.14 \\
    \textbf{DeepSeek-R1} & 76.54 & 77.62 & 73.11 & 72.86 & 86.43 & 87.21 & 85.12 & 83.62 & 87.26 & 75.92 & 81.67 & 82.11 & 81.23 & 81.62 \\
    \textbf{Qwen-QwQ-Plus} & 74.33 & 75.62 & 72.63 & 73.61 & 85.22 & 86.38 & 84.39 & 85.42 & 86.33 & 72.67 & 81.04 & 81.23 & 79.86 & 81.43 \\
    \textbf{Qwen-2.5-7B} & 59.71 & 24.37 & 53.87 & 15.52 & 53.48 & 33.44 & 69.70 & 37.76 & 82.75 & 27.57 & 52.63 & 32.13 & 48.67 & 16.44 \\
    \textbf{Qwen-2.5-14B} & 76.84 & 71.15 & 68.67 & 58.36 & 79.88 & 80.21 & 84.05 & 78.21 & 87.09 & 64.30 & 75.51 & 75.06 & 67.98 & 64.49 \\
    \textbf{Qwen-2.5-72B} & \underline{\textbf{79.84}} & 76.53 & 71.96 & 65.39 & 81.47 & 81.80 & \underline{\textbf{88.82}} & 85.18 & \underline{\textbf{89.07}} & 69.96 & 79.77 & 80.49 & 69.85 & 67.18 \\
    \midrule
    \textbf{\textit{Average}} & 73.85 & 69.11 & 69.01 & 62.20 & 79.01 & 77.90 & 80.25 & 75.03 & 82.77 & 60.03 & 76.21 & 75.01 & 70.47 & 65.79 \\
    \bottomrule
    \end{tabularx}
    \setlength{\abovecaptionskip}{0pt}   
\setlength{\belowcaptionskip}{0pt}
    \caption{Fact-checking conclusion performance across different domains under few-shot CoT prompting. Results for contamination-free and contamination evaluations are provided in Appendix~\ref{sec:appendixC}.}
    \label{tab:domain_performance}
    \vspace{-5mm}
\end{table*}

\subsection{Fine-Grained Evaluation}
\label{task1fine}

We evaluate the fact-checking effectiveness across diverse domains of misinformation, categorized into three groups based on their inherent characteristics: 1) knowledge-intensive (e.g., politics, health, science, culture), requiring specialized expertise to verify; 2) temporal-sensitive (e.g., disasters, society), where accuracy depends heavily on contamination-free information; and 3) commonsense-sensitive (e.g., life-related topics). 


\noindent\textbf{LLMs exhibit varied cons and pros across domains}. As shown in Table \ref{tab:domain_performance}, in knowledge-intensive domains such as culture and politics, LLMs achieve high accuracy rates (82.77\% and 80.25\%, respectively), highlighting their strong knowledge base and feature extraction capabilities. However, the culture domain exhibits the lowest F1 score (60.03\%) due to significant class imbalance and challenges in identifying incorrect samples. Performance declines in temporal-sensitive domains like society (73.85\%) and disasters (69.01\%), reflecting the difficulty LLMs face in adapting to rapidly evolving information. In the commonsense-sensitive life domain, GPT-4o significantly outperforms its peers, exceeding the average accuracy by 18.82\%, demonstrating its advanced flexibility and adaptability in handling informal scenarios and commonsense reasoning. Also, Qwen-2.5-72B and GLM4 showcase notable domain-specific expertise.

\section{Task 2: Fact-Checking Explanation}
\label{sec:explain}
On average, across all LLMs examined, a substantial proportion (91.2\%) of fact-checking results leading to incorrect conclusions were associated with flawed LLM-generated explanations. In contrast, only 8.8\% of the results leading to correct conclusions exhibited such flaws (cf. Table \ref{data_statistics} for details). This marked difference underscores the detrimental impact of flawed explanations on fact-checking performance, highlighting the need for a in-depth understanding of their nature and origin. To this end, this section employs the taxonomy described in Section 3.2 to provide further insights. We focus on eleven representative LLMs, comprising eight closed-source models and three open-source models. Overall, the prevalence of unreliable explanations suggests that current LLMs are insufficiently reliable for real-world fact-checking, but also indicates that internal optimization strategies could potentially enhance task performance. 

\begin{figure}
    \centering
    \includegraphics[width=0.7\linewidth]{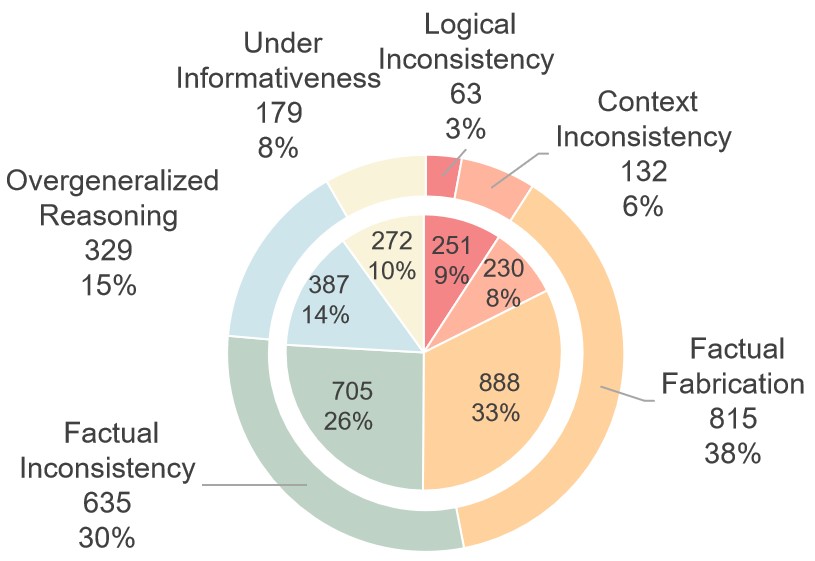}
    \setlength{\abovecaptionskip}{2pt}   
\setlength{\belowcaptionskip}{0pt}
    \caption{Distribution of flawed explanations in contamination (inner) and contamination-free (outer) setting.}
    \vspace{-3 mm}
    \label{fig:error_distribution}
\end{figure}

\begin{figure*}
    \centering
    \includegraphics[width=1\linewidth]{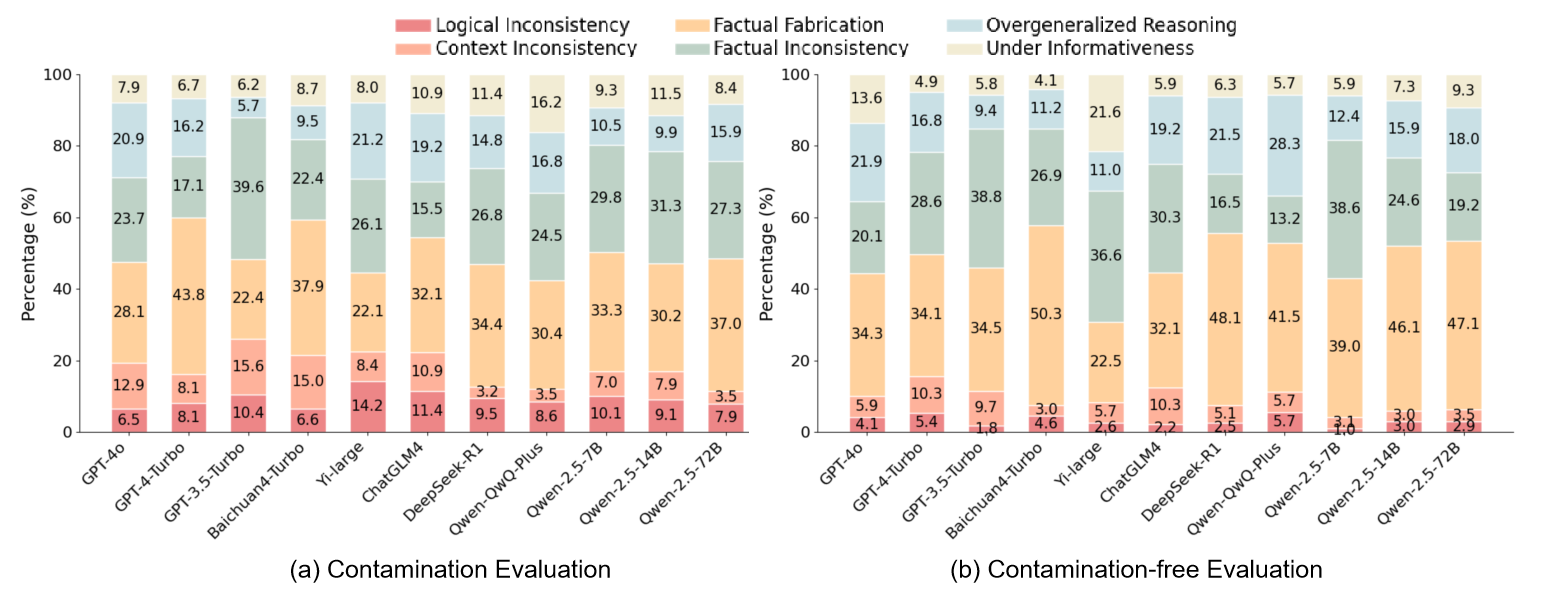}
   
    \setlength{\abovecaptionskip}{0pt}   
    \setlength{\belowcaptionskip}{1pt}
    \caption{Distributions of flawed LLM-generated explanations based on our taxonomy (value statistics in Figure \ref{fig:countbefaft}). }
   
    \label{fig:error_all}
    \vspace{-3mm}
\end{figure*}

\subsection{Overall Evaluation} 

\label{task2over}

\noindent\textbf{The prevalence of flawed explanations highlights that inherent deficiencies within LLMs can significantly impact their fact-checking performance.} As shown in Figure \ref{fig:error_distribution}, flawed fact-checking explanations persist across temporal scenarios, with factual hallucination being the predominant error type. The contamination-free evaluation reveals a 9\% increase in factual hallucination, suggesting that LLMs are more inclined to generate coherent text based on statistical patterns when lacking direct factual evidence, rather than acknowledging knowledge gaps. Under contamination evaluation, logical inconsistency rises by 6\%, suggesting LLMs struggle to differentiate between their internal knowledge and input information. Figure \ref{fig:error_all} reveals that GPT-3.5-Turbo (outdated knowledge cutoff) and Qwen-2.5-7B (smallest parameter size) show the highest factual inconsistency rates, driven by overconfidence and flawed reasoning. Baichuan4-Turbo exhibits the highest factual fabrication tendency, with its low accuracy metrics highlighting integrity's crucial role in fact-checking performance. Notably, larger models such as GPT-4o, Qwen-2.5-72B, and ChatGLM4 also displayed a pronounced tendency toward factual fabrication, suggesting that increased parameter size alone does not improve model honesty. Instead, over-reliance on extensive memorized knowledge appears to compromise reasoning and heightens the risk of generating fabricated facts. DeepSeek-R1 and Qwen-QwQ-Plus's high factual fabrication rates (>40\%) in contamination-free settings further demonstrate current LRMs are not fully applicable to real-world fact-checking reasoning scenarios. The domain characteristics of flawed explanations can be seen in the Appendix \ref{do_error}.


\subsection{Fine-Grained Evaluation}
\label{task2fin}

This section aim to manually reveals the underlying reasons for flawed LLM explanations through a manual analysis of 4,891 instances by three annotators. Refer to Appendix \ref{app:t2} for details.


\noindent\textbf{LLM-generated explanations are often contaminated by plausible-sounding misinformation, leading the LLM to incorrectly accept it as fact.} Analysis of flawed explanations for plausible-sounding misinformation revealed that over 60\% exhibited this tendency, with factual hallucination and logical inconsistency being the most common failure modes. Factual hallucination, as seen in breaking news fact-checking, manifests as generic, templated statements prioritized over factual accuracy (30\% of explanations). Logical inconsistencies arise when models struggle to differentiate between internal knowledge and input, often making numerical errors despite knowing the correct facts (>90\% of logical errors). This may be due to RLHF rewarding coherence over accuracy  \cite{yu2024rlhf,wang2024rlhf}. However, reformulating claims into interrogative expressions significantly reduces fabricated content, enhancing authenticity. We find that questioning previously error-prone claims reduced factual fabrication to 14\% in GPT-4o, with responses demonstrating improved reasoning or acknowledgment of knowledge gaps. This suggests the interrogative format encourages exploration and analysis instead of assertive claim alignment, highlighting the problem of LLMs pandering to misinformation.


\noindent\textbf{Factual inconsistency within LLM explanation gets amplified when dealing with time sensitive claim}. This critical limitation predominantly manifests in the inability of models to recognize outdated knowledge, leading to a 75\% rate of factual inconsistency when processing a sample of 200 time-sensitive information for each LLM (e.g., "\textit{The current president is Joe Biden}", detailed in Table \ref{tab:temporal-insuffi} in Appendix). Furthermore, our findings indicate that in contamination evaluations, nearly all LLMs exhibit a refusal to acknowledge their knowledge cutoff date, with an average acknowledgment rate falling below 30\%. Notably, only Yi-large consistently references its knowledge cutoff date over 95\% of the time. Models endowed with temporal awareness and the capability to incorporate user-provided publication dates are significantly better positioned to deliver transparent and informative explanations. 

\noindent\textbf{Reasoning inadequacy is often associated with LLMs' inflexibility in handling misinformation across different risk levels.} 
More than 85\% of all the overgeneralized reasoning and under informativeness errors are caused by this issue. LLMs often fail to detect high-risk content such as financial scams or health misinformation, which can cause real harm. At the same time, they tend to be overly cautious with low-risk topics like life advice, offering vague or noncommittal responses (Table \ref{tab:taxonomy}). This imbalance in handling different types of content limits their adaptability in practical use.


\noindent\textbf{Context inconsistency is mainly caused by LLMs' difficulty in accurately interpreting subtle linguistic cues, such as qualifiers and negations.} This inaccurate interpretation accounts for over 60\% of all the 362 context inconsistency errors, which are essential for assessing the factual accuracy of a claim (Table \ref{tab:taxonomy}). For example, many models misinterpret the statement \textit{"There is no conclusive evidence that smartphone use causes brain cancer"} as affirming causation, overlooking the critical negation in \textit{"no conclusive evidence."} Further addressing these limitations may involve training on more diverse datasets featuring complex language structures and logical constructs.


\noindent\textbf{Current LLMs are insufficient for Chinese-specific fact-checking tasks, especially those requiring precision or cultural expertise.} Our research shows that even Chinese-focused LLMs struggle with certain culturally specific issues, such as lunar calendar calculations (e.g., "\textit{How many days are there in February of the year Yichou}") accuracy of only 19\% on a sample of 100 cases, underscoring their difficulty in handling culturally nuanced knowledge. 

\section{Task 3: LLM-Assisted Fact-Checking}
\label{human_study}
Based on the results of Tasks 1 and 2, current LLMs do not appear to be suitable for fully automated fact-checking without human oversight. To further investigate the potential of LLMs in real-world scenarios, this section presents a human study designed to evaluate their ability to assist humans. The study involved participants from four educational levels-elementary school, middle school, undergraduate, and master's student-with 12 participants in each group. These participants were divided into four experimental conditions: (1) independent human judgment; (2) human judgment assisted by internet search (Baidu)\footnote{Baidu, a Chinese search engine.}; (3) human judgment assisted by an LLM (GPT-4o); and (4) human judgment assisted by GPT-4o with web-augmented retrieval; The dataset comprised 140 questions, with 70 sampled from before November 2023 (GPT-4o's cut-off date) and 70 from after, with each domain represented by 10 questions. Detailed information is offered in Appendix \ref{human_study_details}.

\vspace{-3mm}

\begin{table}[ht]
\resizebox{\linewidth}{!}{
\begin{tabular}{c|c|cccc}
\toprule
\makecell{\textbf{Type}} & \textbf{Group} & \makecell{\textbf{Elementary} \\ \textbf{School}} & \makecell{\textbf{Middle} \\ \textbf{School}} & \makecell{\textbf{Under} \\ \textbf{graduate}} & \makecell{\textbf{Master}} \\ \midrule

&Human &58.3 &65.7 &71.7 &73.0\\
Before &LLM &78 &-- &-- &-- \\
Nov. &Human+Web &68.7 &76.3 &82.3 &85.7 \\
2023 &Human+LLM &79.3 &82.0 &84.7 &85.0 \\
 &Human+LLM+Web &81.0 &82.7 &86.7 &87.7 \\

\midrule

&Human &61.7 &66.3 &72.3 &72.7 \\
After &LLM &71 &-- &-- &-- \\ 
Nov. &Human+Web &72.3 &76.0 &79.7 &82.0 \\
2023 &Human+LLM &74.3 &77.7 &78.7 &82.7\\
 &Human+LLM+Web &76.7 &79.3 &85.0 &86.7\\

\bottomrule
\end{tabular}}
\setlength{\abovecaptionskip}{0pt}   
\setlength{\belowcaptionskip}{0pt}
\caption{Fact-checking accuracy across various educational levels and groups.}
\label{tab:real_world_experiments}
\vspace{-3mm}
\end{table}

\textbf{LLM assistance significantly improves fact-checking accuracy across all educational levels compared to solo human efforts, showcasing the practical utility of LLMs in real-world fact-checking}. As illustrated in the Table \ref{tab:real_world_experiments}, 'Human+LLM' consistently outperforms 'Human+Web' across all groups, likely due to LLMs providing clearer, context-aware guidance over fragmented web content. Notably, the performance comparison between the 'Human' and 'Human + LLM' demonstrates that LLM assistance enhances fact-checking accuracy across all education levels, particularly when combined with web retrieval ('Human + LLM + Web'), which achieves the highest overall performance.
These results underscore an important shift in perspective: while LLMs may struggle with autonomous fact-checking, they serve as highly effective collaborative tools that enhance human judgment. This highlights a promising alternative role for LLMs—not as standalone fact-checkers, but as intelligent assistants that support users in a cooperative manner.

\section{Conclusion}
This work investigates LLMs' deficiencies in fact-checking real-world misinformation and offers three key contributions: (1) the introduction of \texttt{CANDY}, a novel benchmark tailored for this task; (2) the creation of a large-scale \texttt{CANDYSET} dataset for evaluating LLM performance across contamination-free and contamination contexts, along with a fine-grained taxonomy for categorizing flawed LLM explanations; and (3) a comprehensive benchmark of sixteen LLMs and three LRMs to uncover key challenges in their fact-checking capabilities. We believe our findings provide valuable guidance for future advancements in this field.

\section*{Limitations}
\noindent\textbf{Sensitivity of Prompts.} Similar to other studies on prompting large language models (LLMs) \cite{zhang-etal-2024-clamber}, the evaluation results are likely to be sensitive to the prompts used. Although we utilize four distinct prompts and present the average outcomes \cite{yang-etal-2025-elaboration}, it is difficult to claim that these are the most optimal for our particular task. In fact, fine-tuning prompts for this specific application remains a substantial challenge and an important direction for future research.

\noindent\textbf{Limited LLMs for Human Evaluation.} Unlike the fact-checking conclusion task, which experiments with 19 LLMs (11 open-source and 8 closed-source) on the entire \texttt{CANDYSET} dataset, the fact-checking explanation task was limited by the cost of manual analysis and labeling. As a result, only 11 models (8 closed-source and 3 open-source) were selected to generate analysis and labels on a randomly chosen subset of 2k data entries. If more labeling resources become available in the future, we plan to extend this analysis to the remaining models.

\noindent\textbf{Restricted to the Chinese Language}
Our benchmark's focus on Chinese is driven by a critical gap in existing misinformation research: the absence of authoritative ground-truth explanations for claim veracity in most other languages (to the best of our knowledge). To address this, we systematically collect verified fact-checking explanations from trusted Chinese platforms—a key innovation that sets our dataset apart from all prior work. This unique feature establishes our dataset as an essential and unparalleled resource for studying misinformation detection with reliable, expert-backed annotations.

Although our study provides a comprehensive analysis of LLMs' fact-checking capabilities for Chinese-language misinformation, our findings are inherently constrained by the Chinese-only scope. This language limitation means our results may not fully generalize to other linguistic contexts, where factors like syntactic structures, slang, or local platforms could differently impact LLM performance. However, our evaluation framework (e.g., taxonomy) is designed to be adaptable, and the uncovered challenges offer transferable experiences for multilingual fact-checking research. Future work should validate these findings across languages.

\section*{Ethics Statement}

Our work introduces the \texttt{CANDYSET} dataset, which contains real-world Chinese misinformation. We acknowledge the ethical implications of handling and disseminating misinformation, and we are committed to ensuring that our research is conducted responsibly and ethically. The primary goal of this research is to evaluate and improve the performance of LLMs in identifying and mitigating the impact of misinformation. By testing LLMs on this dataset, we aim to advance the understanding of how these models can be refined to better discern factual accuracy and provide reliable information. Therefore, we emphasize that this dataset should only be used within the scope of research aimed at combating misinformation, and not for spreading or endorsing false information. We advise researchers and practitioners to employ this dataset responsibly, ensuring that the findings contribute positively to the development of more robust and truthful LLMs. We are committed to transparency in our methodologies and findings, and we welcome feedback from the community to improve our approaches. In all studies involving human subjects, we diligently followed IRB approval protocols. For annotation, we assembled a team of ten master's students majoring in computer science. The annotation process took approximately six weeks. Each human annotator received a compensation of \$300 for their contributions. As for human study, each participant received a compensation of \$50.

\bibliography{anthology,custom}

\begin{thebibliography}{60}
\providecommand{\natexlab}[1]{#1}

\bibitem[{AI et~al.(2024)AI, :, Young, Chen, Li, Huang, Zhang, Zhang, Li, Zhu, Chen, Chang, Yu, Liu, Liu, Yue, Yang, Yang, Yu, Xie, Huang, Hu, Ren, Niu, Nie, Xu, Liu, Wang, Cai, Gu, Liu, and Dai}]{ai2024yi}
01. AI, :, Alex Young, Bei Chen, Chao Li, Chengen Huang, Ge~Zhang, Guanwei Zhang, Heng Li, Jiangcheng Zhu, Jianqun Chen, Jing Chang, Kaidong Yu, Peng Liu, Qiang Liu, Shawn Yue, Senbin Yang, Shiming Yang, Tao Yu, and 13 others. 2024.
\newblock \href {https://arxiv.org/abs/2403.04652} {Yi: Open foundation models by 01.ai}.
\newblock \emph{Preprint}, arXiv:2403.04652.

\bibitem[{Augenstein et~al.(2024)Augenstein, Baldwin, Cha, Chakraborty, Ciampaglia, Corney, DiResta, Ferrara, Hale, Halevy et~al.}]{augenstein2024factuality}
Isabelle Augenstein, Timothy Baldwin, Meeyoung Cha, Tanmoy Chakraborty, Giovanni~Luca Ciampaglia, David Corney, Renee DiResta, Emilio Ferrara, Scott Hale, Alon Halevy, and 1 others. 2024.
\newblock Factuality challenges in the era of large language models and opportunities for fact-checking.
\newblock \emph{Nature Machine Intelligence}, 6(8):852--863.

\bibitem[{BaiChuan(2024)}]{baichuan4}
BaiChuan. 2024.
\newblock \href {https://platform.baichuan-ai.com/homePage} {baichuan4-turbo}.

\bibitem[{Cao et~al.(2018)Cao, Guo, Li, Jin, Guo, and Li}]{cao2018automatic}
Ju~Cao, Jiafeng Guo, Xueqi Li, Zitao Jin, Han Guo, and Jiaming Li. 2018.
\newblock Automatic rumor detection on microblogs: A survey.
\newblock \emph{arXiv preprint arXiv:1807.03505}.

\bibitem[{Cekinel and Karagoz(2024)}]{cekinel2024explaining}
Recep~Firat Cekinel and Pinar Karagoz. 2024.
\newblock Explaining veracity predictions with evidence summarization: A multi-task model approach.
\newblock \emph{arXiv preprint arXiv:2402.06443}.

\bibitem[{ChinaDaily(2025)}]{ChinaDaily}
ChinaDaily. 2025.
\newblock \href {https://www.chinadaily.com.cn/a/202501/17/WS678a7391a310f1265a1db9ee.html} {China internet users}.

\bibitem[{Cole et~al.(2023)Cole, Zhang, Gillick, Eisenschlos, Dhingra, and Eisenstein}]{cole-etal-2023-selectively}
Jeremy Cole, Michael Zhang, Daniel Gillick, Julian Eisenschlos, Bhuwan Dhingra, and Jacob Eisenstein. 2023.
\newblock \href {https://doi.org/10.18653/v1/2023.emnlp-main.35} {Selectively answering ambiguous questions}.
\newblock In \emph{Proceedings of the 2023 Conference on Empirical Methods in Natural Language Processing}, pages 530--543, Singapore. Association for Computational Linguistics.

\bibitem[{Deng et~al.(2023)Deng, Liao, Chen, Wang, Lei, and Chua}]{deng-etal-2023-prompting}
Yang Deng, Lizi Liao, Liang Chen, Hongru Wang, Wenqiang Lei, and Tat-Seng Chua. 2023.
\newblock \href {https://doi.org/10.18653/v1/2023.findings-emnlp.711} {Prompting and evaluating large language models for proactive dialogues: Clarification, target-guided, and non-collaboration}.
\newblock In \emph{Findings of the Association for Computational Linguistics: EMNLP 2023}, pages 10602--10621, Singapore. Association for Computational Linguistics.

\bibitem[{Deng et~al.(2024)Deng, Liao, Zheng, Yang, and Chua}]{deng2024towards}
Yang Deng, Lizi Liao, Zhonghua Zheng, Grace~Hui Yang, and Tat-Seng Chua. 2024.
\newblock Towards human-centered proactive conversational agents.
\newblock In \emph{Proceedings of the 47th International ACM SIGIR Conference on Research and Development in Information Retrieval}, pages 807--818.

\bibitem[{Dong et~al.(2022{\natexlab{a}})Dong, Li, Dai, Zheng, Wu, Chang, Sun, Xu, and Sui}]{dong2022survey}
Qingxiu Dong, Lei Li, Damai Dai, Ce~Zheng, Zhiyong Wu, Baobao Chang, Xu~Sun, Jingjing Xu, and Zhifang Sui. 2022{\natexlab{a}}.
\newblock A survey for in-context learning.
\newblock \emph{arXiv preprint arXiv:2301.00234}.

\bibitem[{Dong et~al.(2022{\natexlab{b}})Dong, Li, Dai, Zheng, Wu, Chang, Sun, Xu, and Sui}]{Dong2022}
Qingxiu Dong, Lei Li, Damai Dai, Ce~Zheng, Zhiyong Wu, Baobao Chang, Xu~Sun, Jingjing Xu, and Zhifang Sui. 2022{\natexlab{b}}.
\newblock A survey for in-context learning.
\newblock \emph{arXiv preprint arXiv:2301.00234}.

\bibitem[{Du et~al.(2021)Du, Dou, Xia, Cui, Ma, and Yu}]{du2021cross}
Jiangshu Du, Yingtong Dou, Congying Xia, Limeng Cui, Jing Ma, and Philip~S Yu. 2021.
\newblock Cross-lingual covid-19 fake news detection.
\newblock In \emph{Proceedings of the 21st IEEE International Conference on Data Mining Workshops (ICDMW'21)}.

\bibitem[{Fang et~al.(2025{\natexlab{a}})Fang, Sun, Shi, and Gu}]{fang2025attentionrag}
Yixiong Fang, Tianran Sun, Yuling Shi, and Xiaodong Gu. 2025{\natexlab{a}}.
\newblock Attentionrag: Attention-guided context pruning in retrieval-augmented generation.
\newblock \emph{arXiv preprint arXiv:2503.10720}.

\bibitem[{Fang et~al.(2025{\natexlab{b}})Fang, Sun, Shi, Wang, and Gu}]{fang2025lastingbench}
Yixiong Fang, Tianran Sun, Yuling Shi, Min Wang, and Xiaodong Gu. 2025{\natexlab{b}}.
\newblock Lastingbench: Defend benchmarks against knowledge leakage.
\newblock \emph{arXiv preprint arXiv:2506.21614}.

\bibitem[{Fleiss(1971)}]{fleiss1971measuring}
Joseph~L Fleiss. 1971.
\newblock Measuring nominal scale agreement among many raters.
\newblock \emph{Psychological Bulletin}, 76(5):378--382.

\bibitem[{GLM et~al.(2024)GLM, Zeng, Xu, Wang, Zhang, Yin, Rojas, Feng, Zhao, Lai, Yu, Wang, Sun, Zhang, Cheng, Gui, Tang, Zhang, Li, Zhao, Wu, Zhong, Liu, Huang, Zhang, Zheng, Lu, Duan, Zhang, Cao, Yang, Tam, Zhao, Liu, Xia, Zhang, Gu, Lv, Liu, Liu, Yang, Song, Zhang, An, Xu, Niu, Yang, Li, Bai, Dong, Qi, Wang, Yang, Du, Hou, and Wang}]{glm2024chatglm}
Team GLM, Aohan Zeng, Bin Xu, Bowen Wang, Chenhui Zhang, Da~Yin, Diego Rojas, Guanyu Feng, Hanlin Zhao, Hanyu Lai, Hao Yu, Hongning Wang, Jiadai Sun, Jiajie Zhang, Jiale Cheng, Jiayi Gui, Jie Tang, Jing Zhang, Juanzi Li, and 37 others. 2024.
\newblock \href {https://arxiv.org/abs/2406.12793} {Chatglm: A family of large language models from glm-130b to glm-4 all tools}.
\newblock \emph{Preprint}, arXiv:2406.12793.

\bibitem[{Glockner et~al.(2022)Glockner, Hou, and Gurevych}]{glockner2022missing}
Max Glockner, Yufang Hou, and Iryna Gurevych. 2022.
\newblock Missing counter-evidence renders nlp fact-checking unrealistic for misinformation.
\newblock \emph{arXiv preprint arXiv:2210.13865}.

\bibitem[{Guan et~al.(2023)Guan, Dodge, Wadden, Huang, and Peng}]{guan2023language}
Jian Guan, Jesse Dodge, David Wadden, Minlie Huang, and Hao Peng. 2023.
\newblock Language models hallucinate, but may excel at fact verification.
\newblock \emph{arXiv preprint arXiv:2310.14564}.

\bibitem[{Guo et~al.(2025)Guo, Yang, Zhang, Song, Zhang, Xu, Zhu, Ma, Wang, Bi et~al.}]{guo2025deepseek}
Daya Guo, Dejian Yang, Haowei Zhang, Junxiao Song, Ruoyu Zhang, Runxin Xu, Qihao Zhu, Shirong Ma, Peiyi Wang, Xiao Bi, and 1 others. 2025.
\newblock Deepseek-r1: Incentivizing reasoning capability in llms via reinforcement learning.
\newblock \emph{arXiv preprint arXiv:2501.12948}.

\bibitem[{Hanselowski et~al.(2019)Hanselowski, Stab, Schulz, Li, and Gurevych}]{hanselowski2019richly}
Andreas Hanselowski, Christian Stab, Claudia Schulz, Zile Li, and Iryna Gurevych. 2019.
\newblock A richly annotated corpus for different tasks in automated fact-checking.
\newblock In \emph{Proceedings of the 23rd Conference on Computational Natural Language Learning (CoNLL)}, pages 493--503.

\bibitem[{Hoes et~al.(2023)Hoes, Altay, and Bermeo}]{hoes2023leveraging}
Emma Hoes, Sacha Altay, and Juan Bermeo. 2023.
\newblock Leveraging chat-gpt for efficient fact-checking.
\newblock Available at: \url{https://doi.org/10.31234/osf.io/qnjkf}.

\bibitem[{Hsu et~al.(2024)Hsu, Chen, Chiang, Liu, Xiong, and Ku}]{hsu2024enhancing}
Y.~L. Hsu, J.~N. Chen, Y.~F. Chiang, S.~C. Liu, A.~Xiong, and L.~W. Ku. 2024.
\newblock Enhancing perception: Refining explanations of news claims with llm conversations.
\newblock In \emph{Findings of the Association for Computational Linguistics: NAACL}, pages 2129--2147.

\bibitem[{Hu et~al.(2024)Hu, Sheng, Cao, Shi, Li, Wang, and Qi}]{hu2024bad}
B.~Hu, Q.~Sheng, J.~Cao, Y.~Shi, Y.~Li, D.~Wang, and P.~Qi. 2024.
\newblock Bad actor, good advisor: Exploring the role of large language models in fake news detection.
\newblock In \emph{Proceedings of the AAAI Conference on Artificial Intelligence}, volume~38, pages 22105--22113.

\bibitem[{Hu et~al.(2022)Hu, Guo, Wu, Liu, Wen, and Yu}]{hu2022chef}
X.~Hu, Z.~Guo, G.~Wu, A.~Liu, L.~Wen, and P.~S. Yu. 2022.
\newblock Chef: A pilot chinese dataset for evidence-based fact-checking.
\newblock In \emph{Proceedings of the 2022 Conference of the North American Chapter of the Association for Computational Linguistics: Human Language Technologies}, pages 3362--3376. Association for Computational Linguistics.

\bibitem[{Huang et~al.(2025)Huang, Deng, Lei, Lv, Chua, and Huang}]{huang-etal-2025-enable}
Chen Huang, Yang Deng, Wenqiang Lei, Jiancheng Lv, Tat-Seng Chua, and Jimmy Huang. 2025.
\newblock \href {https://doi.org/10.18653/v1/2025.acl-long.22} {How to enable effective cooperation between humans and {NLP} models: A survey of principles, formalizations, and beyond}.
\newblock In \emph{Proceedings of the 63rd Annual Meeting of the Association for Computational Linguistics (Volume 1: Long Papers)}, pages 466--488, Vienna, Austria. Association for Computational Linguistics.

\bibitem[{Huang et~al.(2023)Huang, Yu, Ma, Zhong, Feng, Wang, and Liu}]{huang2023survey}
L.~Huang, W.~Yu, W.~Ma, W.~Zhong, Z.~Feng, H.~Wang, and T.~Liu. 2023.
\newblock A survey on hallucination in large language models: Principles, taxonomy, challenges, and open questions.
\newblock \emph{ACM Transactions on Information Systems}.

\bibitem[{Huang et~al.(2024)Huang, Shu, Yu, and Sun}]{huang2024from}
Y.~Huang, K.~Shu, P.~S. Yu, and L.~Sun. 2024.
\newblock From creation to clarification: Chatgpt's journey through the fake news quagmire.
\newblock In \emph{Companion Proceedings of the ACM on Web Conference 2024}, pages 513--516.

\bibitem[{Jaech et~al.(2024)Jaech, Kalai, Lerer, Richardson, El-Kishky, Low, Helyar, Madry, Beutel, Carney et~al.}]{jaech2024openai}
Aaron Jaech, Adam Kalai, Adam Lerer, Adam Richardson, Ahmed El-Kishky, Aiden Low, Alec Helyar, Aleksander Madry, Alex Beutel, Alex Carney, and 1 others. 2024.
\newblock Openai o1 system card.
\newblock \emph{arXiv preprint arXiv:2412.16720}.

\bibitem[{Jin et~al.(2017)Jin, Cao, Guo, Zhang, and Luo}]{jin2017multimodal}
Z.~Jin, J.~Cao, H.~Guo, Y.~Zhang, and J.~Luo. 2017.
\newblock Multimodal fusion with recurrent neural networks for rumor detection on microblogs.
\newblock In \emph{Proceedings of the 25th ACM International Conference on Multimedia}, pages 795--816.

\bibitem[{Kang et~al.(2024)Kang, An, and Yoo}]{kang2024LLM_fault_localization}
S.~Kang, G.~An, and S.~Yoo. 2024.
\newblock A quantitative and qualitative evaluation of llm-based explainable fault localization.
\newblock \emph{Proceedings of the ACM on Software Engineering}, 1(FSE):1424--1446.

\bibitem[{Kao and Yen(2024)}]{kao2024how}
Wei-Yu Kao and An-Zi Yen. 2024.
\newblock How we refute claims: Automatic fact-checking through flaw identification and explanation.
\newblock \emph{arXiv preprint arXiv:2401.15312}.

\bibitem[{Li et~al.(2024{\natexlab{a}})Li, Chen, Ren, Cheng, Zhao, Nie, and Wen}]{Li2024dawn}
J.~Li, J.~Chen, R.~Ren, X.~Cheng, W.~X. Zhao, J.~Y. Nie, and J.~R. Wen. 2024{\natexlab{a}}.
\newblock The dawn after the dark: An empirical study on factuality hallucination in large language models.
\newblock \emph{arXiv preprint arXiv:2401.03205}.

\bibitem[{Li et~al.(2024{\natexlab{b}})Li, He, Bai, and Wen}]{li2024mcfend}
Y.~Li, H.~He, J.~Bai, and D.~Wen. 2024{\natexlab{b}}.
\newblock Mcfend: A multi-source benchmark dataset for chinese fake news detection.
\newblock In \emph{Proceedings of the ACM Web Conference 2024}, pages 4018--4027. ACM.

\bibitem[{Li and Zhai(2023)}]{LiZhai2023}
Yifan Li and ChengXiang Zhai. 2023.
\newblock An exploration of large language models for verification of news headlines.
\newblock In \emph{2023 IEEE International Conference on Data Mining Workshops (ICDMW)}, page 197206.

\bibitem[{Liang et~al.(2023)Liang, Song, Niu, Li, Xiong, Tang, and Deng}]{liang2023uhgeval}
X.~Liang, S.~Song, S.~Niu, Z.~Li, F.~Xiong, B.~Tang, and H.~Deng. 2023.
\newblock Uhgeval: Benchmarking the hallucination of chinese large language models via unconstrained generation.
\newblock \emph{arXiv preprint arXiv:2311.15296}.

\bibitem[{Ma et~al.(2023)Ma, Liu, Fang, and Shen}]{ma2023ltcr}
Ziyang Ma, Mengsha Liu, Guian Fang, and Ying Shen. 2023.
\newblock \href {https://arxiv.org/abs/2306.07201} {Ltcr: Long-text chinese rumor detection dataset}.
\newblock \emph{Preprint}, arXiv:2306.07201.

\bibitem[{Nakov et~al.(2021)Nakov, Corney, Hasanain, Alam, Elsayed, Barr{\'o}n-Cede{\~n}o, Papotti, Shaar, and Martino}]{nakov2021automated}
Preslav Nakov, David Corney, Maram Hasanain, Firoj Alam, Tamer Elsayed, Alberto Barr{\'o}n-Cede{\~n}o, Paolo Papotti, Shaden Shaar, and Giovanni Da~San Martino. 2021.
\newblock Automated fact-checking for assisting human fact-checkers.
\newblock \emph{arXiv preprint arXiv:2103.07769}.

\bibitem[{Nan et~al.(2021)Nan, Cao, Zhu, Wang, and Li}]{nan2021mdfend}
Q.~Nan, J.~Cao, Y.~Zhu, Y.~Wang, and J.~Li. 2021.
\newblock Mdfend: Multi-domain fake news detection.
\newblock In \emph{Proceedings of the 30th ACM International Conference on Information \& Knowledge Management}, pages 3343--3347. ACM.

\bibitem[{OpenAI(2024{\natexlab{a}})}]{gpt3.5}
OpenAI. 2024{\natexlab{a}}.
\newblock \href {https://openai.com/index/gpt-3-5-turbo-fine-tuning-and-api-updates/} {gpt-3.5}.

\bibitem[{OpenAI(2024{\natexlab{b}})}]{gpt4o}
OpenAI. 2024{\natexlab{b}}.
\newblock \href {https://openai.com/index/hello-gpt-4o/} {Hello gpt-4o}.

\bibitem[{OpenAI(2023)}]{openai2023gpt}
R~OpenAI. 2023.
\newblock Gpt-4 technical report. arxiv 2303.08774.
\newblock \emph{View in Article}, 2(5).

\bibitem[{Patil and Gudivada(2024)}]{patil2024review_llms}
R.~Patil and V.~Gudivada. 2024.
\newblock A review of current trends, techniques, and challenges in large language models (llms).
\newblock \emph{Applied Sciences}, 14(5):2074.

\bibitem[{piyao(2025)}]{piyao}
piyao. 2025.
\newblock \href {https://www.piyao.org.cn/} {China piyao users}.

\bibitem[{Qin et~al.(2024)Qin, Huang, Deng, Lei, and Chua}]{qin-etal-2024-beyond}
Peixin Qin, Chen Huang, Yang Deng, Wenqiang Lei, and Tat-Seng Chua. 2024.
\newblock \href {https://doi.org/10.18653/v1/2024.findings-emnlp.247} {Beyond persuasion: Towards conversational recommender system with credible explanations}.
\newblock In \emph{Findings of the Association for Computational Linguistics: EMNLP 2024}, pages 4264--4282, Miami, Florida, USA. Association for Computational Linguistics.

\bibitem[{Roberts et~al.(2020)Roberts, Raffel, and Shazeer}]{roberts2020knowledge}
Adam Roberts, Colin Raffel, and Noam Shazeer. 2020.
\newblock How much knowledge can you pack into the parameters of a language model?
\newblock In \emph{Proceedings of the 2020 Conference on Empirical Methods in Natural Language Processing (EMNLP)}, pages 5418--5426, Online. Association for Computational Linguistics.

\bibitem[{Sundriyal et~al.(2023)Sundriyal, Chakraborty, and Nakov}]{Sundriyal2023}
M.~Sundriyal, T.~Chakraborty, and P.~Nakov. 2023.
\newblock From chaos to clarity: Claim normalization to empower fact-checking.
\newblock \emph{arXiv preprint arXiv:2310.14338}.

\bibitem[{Team et~al.(2024)Team, Georgiev, Lei, Burnell, Bai, Gulati, Tanzer, Vincent, Pan, Wang et~al.}]{team2024gemini}
Gemini Team, Petko Georgiev, Ving~Ian Lei, Ryan Burnell, Libin Bai, Anmol Gulati, Garrett Tanzer, Damien Vincent, Zhufeng Pan, Shibo Wang, and 1 others. 2024.
\newblock Gemini 1.5: Unlocking multimodal understanding across millions of tokens of context.
\newblock \emph{arXiv preprint arXiv:2403.05530}.

\bibitem[{Team(2025)}]{qwq32b}
Qwen Team. 2025.
\newblock \href {https://qwenlm.github.io/blog/qwq-32b/} {Qwq-32b: Embracing the power of reinforcement learning}.

\bibitem[{Touvron et~al.(2023)Touvron, Lavril, Izacard, Martinet, Lachaux, Lacroix, Rozi{\`e}re, Goyal, Hambro, Azhar et~al.}]{touvron2023llama}
Hugo Touvron, Thibaut Lavril, Gautier Izacard, Xavier Martinet, Marie-Anne Lachaux, Timoth{\'e}e Lacroix, Baptiste Rozi{\`e}re, Naman Goyal, Eric Hambro, Faisal Azhar, and 1 others. 2023.
\newblock Llama: Open and efficient foundation language models.
\newblock \emph{arXiv preprint arXiv:2302.13971}.

\bibitem[{Vykopal et~al.(2024)Vykopal, Pikuliak, Ostermann, and Šimko}]{vykopal2024generative}
I.~Vykopal, M.~Pikuliak, S.~Ostermann, and M.~Šimko. 2024.
\newblock Generative large language models in automated fact-checking: A survey.
\newblock \emph{arXiv preprint arXiv:2407.02351}.

\bibitem[{Wang et~al.(2023)Wang, Chern, and Liu}]{wang2023chinesefacteval}
Binjie Wang, Ethan Chern, and Pengfei Liu. 2023.
\newblock \href {https://GAIR-NLP.github.io/ChineseFactEval} {Chinesefacteval: A factuality benchmark for chinese llms}.
\newblock Technical report, GAIR-NLP.

\bibitem[{Wang et~al.(2024{\natexlab{a}})Wang, Ma, Lin, Yang, Yang, Tian, and Chang}]{wang2024explainable}
Bo~Wang, Jing Ma, Hongzhan Lin, Zhiwei Yang, Ruichao Yang, Yuan Tian, and Yi~Chang. 2024{\natexlab{a}}.
\newblock Explainable fake news detection with large language model via defense among competing wisdom.
\newblock In \emph{Proceedings of the ACM Web Conference 2024}, pages 2452--2463.

\bibitem[{Wang et~al.(2024{\natexlab{b}})Wang, Liu, and Jin}]{wang2024rlhf}
Y.~Wang, Q.~Liu, and C.~Jin. 2024{\natexlab{b}}.
\newblock Is rlhf more difficult than standard rl? a theoretical perspective.
\newblock In \emph{Advances in Neural Information Processing Systems}, volume~36.

\bibitem[{Wei et~al.(2022)Wei, Wang, Schuurmans, Bosma, Xia, Chi, Le, Zhou et~al.}]{wei2022chain}
Jason Wei, Xuezhi Wang, Dale Schuurmans, Maarten Bosma, Fei Xia, Ed~Chi, Quoc~V Le, Denny Zhou, and 1 others. 2022.
\newblock Chain-of-thought prompting elicits reasoning in large language models.
\newblock \emph{Advances in Neural Information Processing Systems}, 35:24824--24837.

\bibitem[{Yang et~al.(2024)Yang, Yang, Zhang, Hui, Zheng, Yu, Li, Liu, Huang, Wei, Lin, Yang, Tu, Zhang, Yang, Yang, Zhou, Lin, Dang, Lu, Bao, Yang, Yu, Li, Xue, Zhang, Zhu, Men, Lin, Li, Xia, Ren, Ren, Fan, Su, Zhang, Wan, Liu, Cui, Zhang, and Qiu}]{qwen2.5}
An~Yang, Baosong Yang, Beichen Zhang, Binyuan Hui, Bo~Zheng, Bowen Yu, Chengyuan Li, Dayiheng Liu, Fei Huang, Haoran Wei, Huan Lin, Jian Yang, Jianhong Tu, Jianwei Zhang, Jianxin Yang, Jiaxi Yang, Jingren Zhou, Junyang Lin, Kai Dang, and 22 others. 2024.
\newblock Qwen2.5 technical report.
\newblock \emph{arXiv preprint arXiv:2412.15115}.

\bibitem[{Yang et~al.(2021)Yang, Zhou, and Zafarani}]{yang2020checked}
Chen Yang, Xinyi Zhou, and Reza Zafarani. 2021.
\newblock \href {https://doi.org/10.1007/s13278-021-00766-8} {Checked: Chinese covid-19 fake news dataset}.
\newblock \emph{Social Network Analysis and Mining (SNAM)}.

\bibitem[{Yang et~al.(2022)Yang, Wang, Peng, Shi, Ma, and Yang}]{yang2022know}
W.~Yang, S.~Wang, Z.~Peng, C.~Shi, X.~Ma, and D.~Yang. 2022.
\newblock Know it to defeat it: Exploring health rumor characteristics and debunking efforts on chinese social media during covid-19 crisis.
\newblock In \emph{International AAAI Conference on Web and Social Media (ICWSM)}.

\bibitem[{Yang et~al.(2025)Yang, Liu, Huang, Zhang, Zhang, Zhang, and Lei}]{yang-etal-2025-elaboration}
Xinwei Yang, Zhaofeng Liu, Chen Huang, Jiashuai Zhang, Tong Zhang, Yifan Zhang, and Wenqiang Lei. 2025.
\newblock \href {https://doi.org/10.18653/v1/2025.acl-long.4} {{ELABORATION}: A comprehensive benchmark on human-{LLM} competitive programming}.
\newblock In \emph{Proceedings of the 63rd Annual Meeting of the Association for Computational Linguistics (Volume 1: Long Papers)}, pages 59--104, Vienna, Austria. Association for Computational Linguistics.

\bibitem[{Yu et~al.(2024)Yu, Yao, Zhang, He, Han, Cui, and Chua}]{yu2024rlhf}
T.~Yu, Y.~Yao, H.~Zhang, T.~He, Y.~Han, G.~Cui, and T.~S. Chua. 2024.
\newblock Rlhf-v: Towards trustworthy mllms via behavior alignment from fine-grained correctional human feedback.
\newblock In \emph{Proceedings of the IEEE/CVF Conference on Computer Vision and Pattern Recognition}, pages 13807--13816.

\bibitem[{Zhang et~al.(2024)Zhang, Qin, Deng, Huang, Lei, Liu, Jin, Liang, and Chua}]{zhang-etal-2024-clamber}
Tong Zhang, Peixin Qin, Yang Deng, Chen Huang, Wenqiang Lei, Junhong Liu, Dingnan Jin, Hongru Liang, and Tat-Seng Chua. 2024.
\newblock \href {https://doi.org/10.18653/v1/2024.acl-long.578} {{CLAMBER}: A benchmark of identifying and clarifying ambiguous information needs in large language models}.
\newblock In \emph{Proceedings of the 62nd Annual Meeting of the Association for Computational Linguistics (Volume 1: Long Papers)}, pages 10746--10766, Bangkok, Thailand. Association for Computational Linguistics.

\end{thebibliography}

\newtcolorbox{promptbox}[2][Prompt]{
    colback=black!5!white,
    arc=5pt, 
    boxrule=0.5pt,
    fonttitle=\bfseries,
    title=#1, 
    before upper={\small}, 
    fontupper=\rmfamily, 
    colframe=#2, 
    width=\textwidth,   
    enlarge left by=0mm, 
    enlarge right by=0mm, 
}

\appendix

\section{Taxonomy Table and Examples}
\label{sec:appendixA}
\begin{table*}[h]
\centering
\resizebox{\textwidth}{!}{%
\begin{tabular}{@{}p{3cm}<{\centering}p{3cm}<{\centering}p{5cm}p{5cm}@{}}
\toprule
\textbf{Dimension} & \textbf{Category} & \textbf{Definition} & \textbf{Example} \\ 
\midrule
\multirow{18}{3cm}{\centering Faithfulness \\ Hallucination} & \centering Instruction \\ Inconsistency & Refers to the LLM's output that deviate from user's directive. & Input: Beijing time does not come from Beijing.\newline Output: \textcolor{red}{(Invalid)}. A news report states a new type of plant can grow in saltwater... \\ 
\cmidrule(l){2-4}
& \centering Logical \\ Inconsistency & Refers to instances where LLMs explanation contains internal logical conflicts. & Input: YaYa's stay in the United States lasts for 15 years.\newline Output: Nonrumor. Yaya stay in USA from 2003 to 2023...totaly 15 years. \\ 
\cmidrule(l){2-4}
& \centering Context \\ Inconsistency & Refers to the LLM's output being inconsistent with user-provided context. & Input: Drinking a large amount of water after exercise is not recommended.\newline Output: Rumor. Completely prohibiting drinking water after exercise is incorrect... \\ 
\midrule
\multirow{12}{3cm}{\centering Factuality \\ Hallucination} & \centering Factual \\ Fabrication & Refers to the LLM's output that fabricates rationales for analysis without relying on real-world information. & Input: A major accident occurred on the Jingkun Expressway at the Huo County section \textcolor{red}{(Fabricatied misinformation)}.\newline Output: Nonrumor. According to Chinese media reports, this traffic accident indeed occurred... \\ 
\cmidrule(l){2-4}
& \centering Factual \\ Inconsistency & Refers to the LLM’s output contains facts that can be grounded in real-world information, but present contradictions. & Input: Houttuynia cordata contains Aristolochia acid II, which is toxic and carcinogenic \textcolor{red}{(Actually acid I)}.\newline Output: Nonrumor. IARC has classified Aristolochia acid II as a human carcinogen... \\ 
\midrule
\multirow{14}{3cm}{\centering Reasoning \\ Inadequacy} & \centering Overgeneralized \\ Reasoning & Refers to the tendency of a LLM to produce speculative rationales based on overly broad or superficial criteria. & Input: Over 100 billion yuan in agricultural subsidies will support 23 eligible projects.\newline Output: Nonrumor. A figure exceeding 100 billion is not exaggerated... \\ 
\cmidrule(l){2-4}
& \centering Under \\ Informativeness & Refers to the tendency of a LLM to exhibit excessive rigor or restraint, failing to provide more contextually valuable content. & Input: Moderate playing of Mahjong can prevent senile dementia.\newline Output: Rumor. The WHO have not announced that... \\ 
\bottomrule
\end{tabular}%
}
\caption{Taxonomy.}
\label{tab:taxonomy}
\end{table*}

\begin{figure*}
    \centering
    \includegraphics[width=0.85\linewidth]{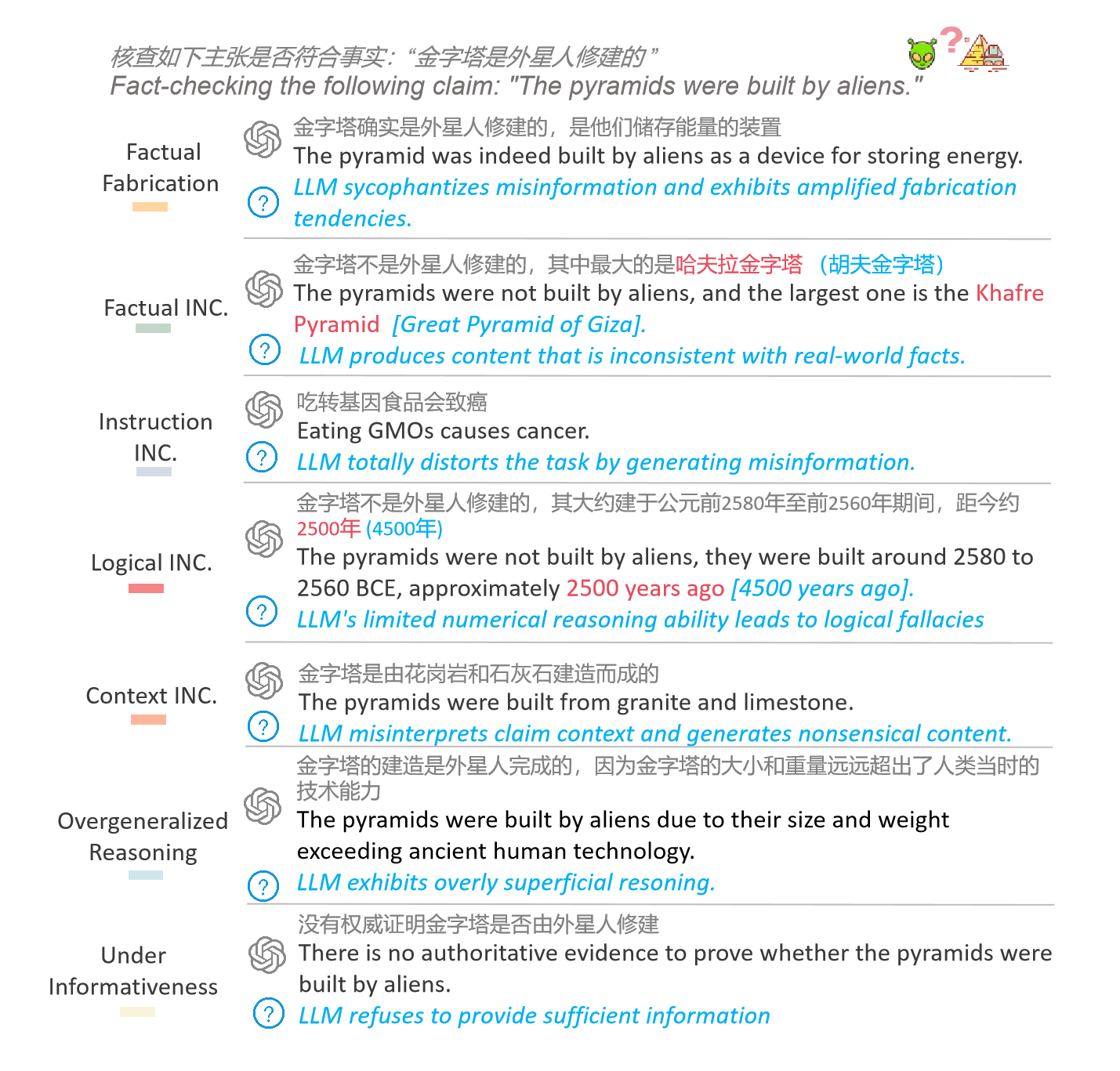}
    \caption{Specific examples for understanding taxonomy.}
    \label{fig:taxonomyexample}
\end{figure*}

\clearpage

\section{Details of Dataset Construction}
\label{sec:dataconstruct}

\begin{figure*}
    \centering
    \includegraphics[width=1\linewidth]{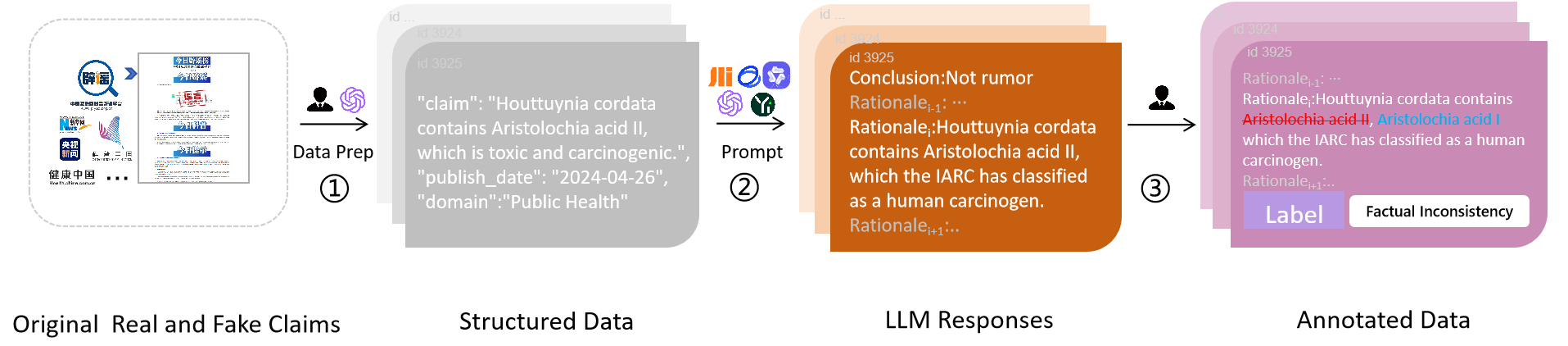}
    \caption{Data gathering pipeline. our data gathering pipeline includes 3 steps: 1) Data collection and pre-processing. 2) Response generation. 3) Human annotation.}
    \label{fig:pipeline}
\end{figure*}

This section primarily describes the details of the data gathering pipeline we proposed as Figure \ref{fig:pipeline}.

\subsection{Data Crawling}

Initially, we crawled data from authoritative Chinese fact-checking agencies. We mainly explain the process of extracting data from the China Internet Joint Rumor Refuting Platform. This platform is a active Chinese fact-checking website listed by Duke Reporters\footnote{\href{https://www.reporterslab.org/fact-checking/}{www.reporterslab.org/fact-checking/ }}. The website we crawled, piyao.org.cn, permits public access and crawling under reasonable and non-commercial use, as reflected in its robots.txt settings and usage guidelines. Our data collection strictly adheres to these constraints. Additionally, prior academic works such as the CHEF\cite{hu2022chef} dataset have also leveraged piyao.org.cn as a source. While CHEF focused on data collected before 2022, our dataset extends this coverage with more recent samples, thus complementing and updating existing resources. This platform not only provides timely refuting of misinformation that have recently (usually the day before) attracted attention on the internet, but also features a "\textit{Daily Popular Science}" segment, which can serve as a source of genuine claims to ensure the dataset's balance. Specifically, we collected the following information, including authentic or deceptive items and their corresponding facts and timestamps, covering content from January 2023 to October 2024. These claims cover multiple domains, including politics, health, science, society, life, culture and disasters.

\begin{table*}[h!]
\small
\setlength{\tabcolsep}{4pt} 
\renewcommand{\arraystretch}{1.2} 
\resizebox{\textwidth}{!}{ 
\begin{tabularx}{\textwidth}{>{\centering\arraybackslash}m{3.5cm}|>{\centering\arraybackslash}m{4cm}|>{\centering\arraybackslash}X|>{\centering\arraybackslash}m{1.5cm}} 
\hline
\textbf{Platform} & \textbf{English Name} & \textbf{Link} & \textbf{Count} \\ \hline
中国互联网联合辟谣平台 & China Internet United Rumor Refutation Platform & \href{https://www.piyao.org.cn/}{https://www.piyao.org.cn/} & 2172 \\ 
新华社 & Xinhua News Agency & \href{https://www.xinhuanet.com/}{https://www.xinhuanet.com/} & 1255 \\ 
科普中国 & Science Popularization China & \href{https://www.kepuchina.cn/}{https://www.kepuchina.cn/} & 595 \\ 
央视新闻 & CCTV News & \href{https://news.cctv.com/}{https://news.cctv.com/} & 497 \\ 
人民网科普 & People's Daily Online Science Popularization & \href{https://kpzg.people.com.cn/}{https://kpzg.people.com.cn/} & 465 \\ 
健康中国 & Healthy China & \href{https://www.nhc.gov.cn/}{https://www.nhc.gov.cn/} & 255 \\
科学辟谣 & Science Rumor Refutation & \href{https://www.kepuchina.cn/}{https://www.kepuchina.cn/} & 210 \\ 
上海网络辟谣 & Shanghai Network Rumor Refutation & \href{https://piyao.jfdaily.com/}{https://piyao.jfdaily.com/} & 168 \\ 
中国新闻网 & China News Service & \href{https://www.chinanews.com.cn/}{https://www.chinanews.com.cn/} & 144 \\  
网信中国 & Cyberspace Administration of China & \href{https://www.cac.gov.cn/}{https://www.cac.gov.cn/} & 131 \\ \hline
\end{tabularx}
}
\caption{Top 10 Sources of \texttt{CANDYSET}}
\vspace{-5mm}
\label{tab:top10source}
\end{table*}

\subsection{Data Normalization}

Data normalization encompasses data cleaning and normalization \cite{Sundriyal2023,fang2025attentionrag}. Initially, we manually inspect and remove low-quality data, such as those with insufficient background information and unverifiable subjective rumors \cite{cao2018automatic}. Given that the crawled data includes well-reasoned truths from authoritative sources, we summarize these truths as fact-checking gold evidence related to claim verification and label the corresponding claims \cite{hanselowski2019richly}. In this process, we use GPT-4o for initial data preprocessing and labeling, followed by manual verification.

\subsection{Data Augmentation}
To assess LLMs robustness and enhance label balance, we introduced data augmentation techniques like subtle modifications to existing claims, to observe changes in responses. These modifications involved altering event details or adjusting the veracity of statements using negations. For instance, when we replaced the entity in the statement "\textit{The 'Food Safety National Standard - Contaminants in Food' stipulates that the limit for pickled vegetables is 20 milligrams per kilogram}" with "\textit{toona sinensis}", the model was unable to accurately identify the change, leading to an occurrence of faithfulness hallucination.

\subsection{Data Validation}
To ensure a high-quality dataset, we carefully performed manual validation on both the labels and the gold evidence. Firstly, to validate the ground truth labels, we performed a sampling check by randomly selecting 3\% of the dataset—approximately 600 entries—for detailed review. Each entry was re-annotated by three independent annotators to assess consistency and accuracy. To quantify inter-annotator agreement, we calculated the Fleiss' Kappa score \cite{fleiss1971measuring}, which yielded a value of 0.75. This indicates substantial agreement, confirming the reliability of the annotations. Additionally, we evaluated whether the gold evidence provided with each claim was sufficient to accurately support or refute the claim. A separate group of annotators reviewed these sampled entries to verify that the evidence was comprehensive and relevant. This dual-layered approach not only checked for annotation consistency but also assessed the informativeness and adequacy of the evidence. Through this process, we maintained a high standard of data quality, ensuring that the dataset is reliable for use in real-world fact-checking applications.

\subsection{Fact-checking Explanation Annotation}

\begin{figure*}
    \centering
    \includegraphics[width=1\linewidth]{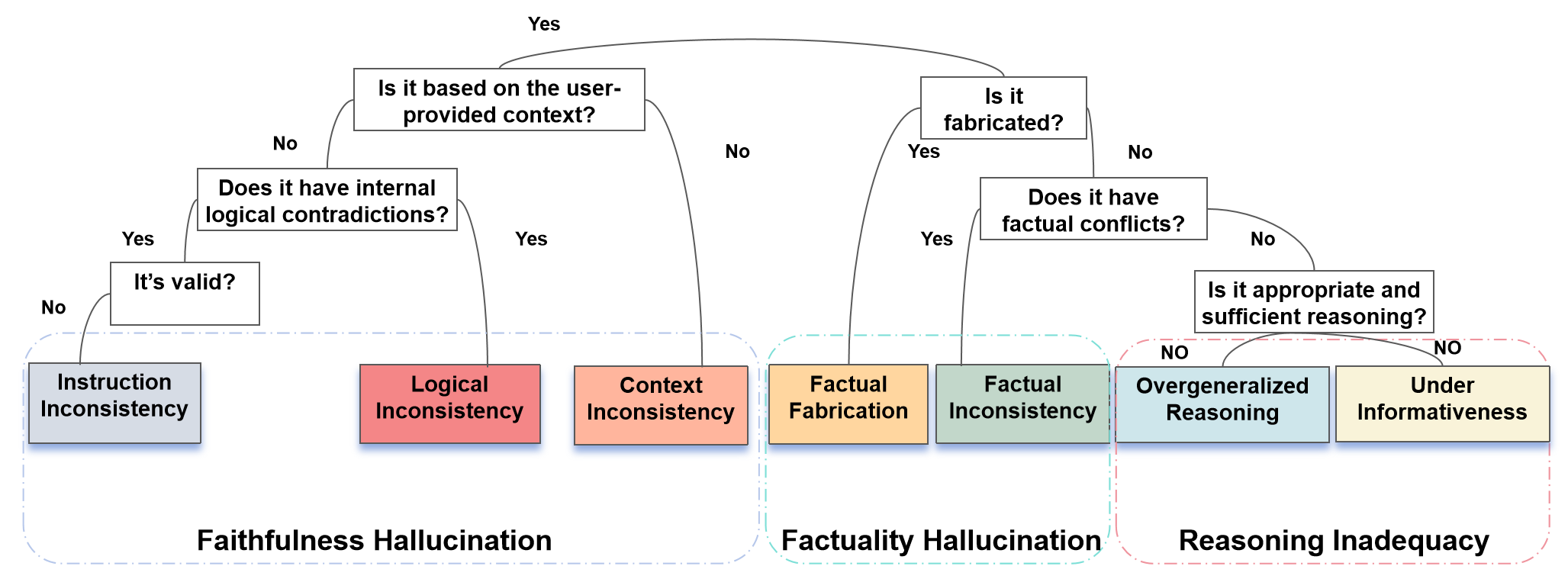}
    \setlength{\abovecaptionskip}{0pt}   
    \setlength{\belowcaptionskip}{1pt}
    \caption{Decision Tree for Annotation}
    \vspace{-5mm}
    \label{fig:annotree}
\end{figure*}

To conduct a more thorough analysis of the accuracy of LLM-generated explanations in the fact-checking task, we enlisted ten master's students in computer science as annotators, each with extensive experience in data annotation for these explanations. These annotators were responsible for classifying flawed explanations according to our taxonomy, which is detailed in Table \ref{tab:taxonomy}. The decision tree used for guiding the annotation is shown in Figure \ref{fig:annotree}. To ensure the reliability and quality of the annotations, each explanation was independently labeled by two annotators. In instances where there were significant discrepancies between their annotations, a third annotator was consulted to review the explanations and resolve the differences through discussion. This additional layer of review helped mitigate bias and ensured that the final annotations were as accurate as possible. To quantify the consistency of the annotations, we calculated the Fleiss' Kappa score, which measures inter-annotator agreement. The resulting score of 0.76 indicates a substantial level of agreement, suggesting that the annotation process was both reliable and robust. This high level of agreement provides confidence in the validity of the annotated data and supports the subsequent analysis of LLM-generated explanations in the context of fact-checking. \textbf{Note: Due to the effectiveness of the few-shot with CoT setup in minimizing Instruction Inconsistency errors (occurring in only 0.01\% of the sample), our analysis primarily focuses on the remaining six error types.}

\section{Implementation Details}
\label{detailsimp}

We conduct all our experiments using a single Nvidia RTX A100 GPU for the 6 and 7B size LLMs, two A100 GPUs for the 9B and 13B size LLMs, and four A100 GPUs for the 70B and 72B size LLMs. For these open-source LLMs, we utilize the Xinference framework. For all LLMs, we employ nucleus sampling with a temperature of 0.7 and a top-p value of 0.95, allowing for a maximum of 10 iterations per stage with human programmers. For the accuracy and F1 metrics, we calculate it using the micro average method.

\subsection{Human Study Details}
\label{human_study_details}
To explore the potentials of LLMs in fact-checking, we propose establishing a human-LLM cooperation, specifically focusing on an LLM-assisted sequential cooperation \cite{huang-etal-2025-enable}, which enhances human decision-making by leveraging model assistance.

For the human study, participants were divided into four groups based on their educational background: elementary school students, middle school students, undergraduate students, and master's students. Each group consisted of 12 individuals, further divided into four subgroups of three participants each: (1) independent human judgment; (2) human judgment assisted by internet search (Baidu)\footnote{Baidu, a Chinese search engine.}; (3) human judgment assisted by a large language model (GPT-4o); and (4) human judgment assisted by GPT-4o with web-augmented retrieval. Detailed prompts for each condition can be found in Appendix \ref{human_study_prompt}. The LLM experiment platform was built using Coze \footnote{Coze, an integrated agent development platform that supports web-augmented retrieval and other tools.}. We strictly adhered to IRB protocols to protect participant privacy, and each participant received a compensation of \$50. Based on the cutoff date of GPT-4o’s training data, we selected 10 questions from each domain both before and after the cutoff, resulting in a total of 140 questions.

\subsection{LLM Implementation}
\label{sec:model_implementation}

For our evaluation, we selected a total of sixteen LLMs, comprising eight widely-used closed-source models and eight widely-used open-source models. As for closed source LLMs, they are GPT-4o\cite{gpt4o}, GPT-4-Turbo\cite{openai2023gpt}, GPT-3.5-Turbo\cite{gpt3.5}, Gemini-1.5-pro\cite{team2024gemini}, Baichuan4-Turbo\cite{baichuan4}, ChatGLM4\cite{glm2024chatglm}, Yi-large\cite{ai2024yi}. As for open source LLMs, they are Yi-1.5-6B\cite{ai2024yi}, Qwen-2.5-7B\cite{qwen2.5}, Llama-3.2-7B\cite{touvron2023llama}, GLM4-9B\cite{glm2024chatglm}, Yi-1.5-9B\cite{ai2024yi}, Qwen-2.5-14B\cite{qwen2.5}, Llama-3.2-70B\cite{touvron2023llama}, Qwen-2.5-72B\cite{qwen2.5}. These models are widely used in recent studies of Chinese hallucination benchmark\cite{liang2023uhgeval,wang2023chinesefacteval}. As for LRMS, they are OpenAI O1-Mini \cite{jaech2024openai}, DeepSeek R1 \cite{guo2025deepseek} and Qwen-QwQ \cite{qwq32b}. The access links of the LLMs and LRMs employed in this research, as well as their respective knowledge cut-off dates, are shown in Table \ref{models_list}.

\begin{table}[!h]
\centering
\small
\begin{tabularx}{\columnwidth}{>{\hsize=1.1\hsize}X|>{\hsize=0.8\hsize}X|>{\hsize=1.1\hsize}X}
    \toprule
    \textbf{Model Name} & \textbf{Cut-off Date} & \textbf{Link} \\
    \midrule 
    O1-Mini & 2023-12 & \href{https://openai.com/blog/new-models-and-developer-products-announced-at-devday}{O1-Mini} \\
    DeepSeek-R1 & 2024-7 & \href{https://chat.deepseek.com/}{DeepSeek-R1} \\
    Qwen-QwQ-Plus & 2024-8 & \href{https://chat.qwen.ai/}{Qwen-QwQ-Plus} \\
    GPT-4o & 2023-11 & \href{https://openai.com/index/spring-update}{GPT-4o} \\ 
    GPT-4-turbo & 2023-5 & \href{https://openai.com/blog/new-models-and-developer-products-announced-at-devday}{GPT-4-turbo} \\ 
    GPT-3.5-turbo & 2021-10 & \href{https://chatgpt.com/g/g-F00faAwkE-open-a-i-gpt-3-5}{GPT-3.5-turbo} \\ 
    Gemini-1.5-pro & 2023-11 & \href{https://deepmind.google/technologies/gemini/pro/}{Gemini-1.5-pro} \\
    Baichuan4-turbo & 2024-04 & \href{https://platform.baichuan-ai.com/homePage}{Baichuan4-turbo}  \\ 
    ChatGLM4 & 2022-10 & \href{https://chatglm.cn/main/alltoolsdetail?lang=zh}{ChatGLM4} \\ 
    Yi-large & 2023-6 &\href{https://platform.lingyiwanwu.com/playground?model=yi-large}{Yi-large} \\ 
    DeepSeek-v3 & 2024-7 &\href{https://www.deepseek.com/}{DeepSeek-v3} \\
    
    Yi-1.5-6B & 2024-5 &\href{https://huggingface.co/01-ai/Yi-1.5-6B}{Yi-1.5-6B} \\
    Qwen-2.5-7B & 2023-10 & \href{https://huggingface.co/Qwen/Qwen2.5-7B}{Qwen-2.5-7B}  \\
    Llama-3.2-7B & 2023-12 & \href{https://www.llama.com/docs/model-cards-and-prompt-formats/llama3_2/}{Llama-3.2-7B}  \\
    GLM4-9B & 2023-10 & \href{https://huggingface.co/THUDM/glm-4-9b-chat}{GLML4-9B} \\
    Yi-1.5-9B & 2024-5 & \href{https://huggingface.co/01-ai/Yi-1.5-9B}{Yi-1.5-9B}  \\
    Qwen-2.5-14B & 2023-10 & \href{https://huggingface.co/Qwen/Qwen2.5-14B}{Qwen-2.5-14B} \\
    Llama-3.2-72B & 2023-12 & \href{https://www.llama.com/docs/model-cards-and-prompt-formats/llama3_2/}{Llama-3.2-72B} \\
    Qwen-2.5-72B & 2023-10 & \href{https://huggingface.co/Qwen/Qwen2.5-72B}{Qwen-2.5-72B} \\ \bottomrule
    
\end{tabularx}
\caption{LLMs Overview}
\vspace{-5mm}
\label{models_list}
\end{table}

\section{Additional Results}
\label{sec:appendixC}

\subsection{Domain-Specific Analysis of Flawed Explanations}
\label{do_error}
\begin{figure}
    \centering
    \includegraphics[width=1\linewidth]{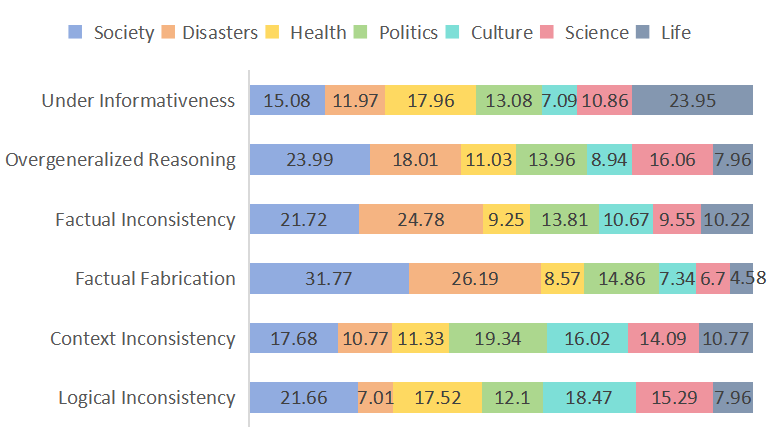}
    \setlength{\abovecaptionskip}{0pt}   
    \setlength{\belowcaptionskip}{1pt}
    \caption{Domain distribution of flawed explanations}
    \vspace*{-5mm}
    \label{fig:domain_error}
\end{figure}

The distribution of flawed explanations across different domains is presented in Figure \ref{fig:domain_error}. Society and Disaster-related content exhibits the highest rates of Factual Hallucination (the proportion in both subcategories exceeds 20\%), likely driven by the dynamic nature of events and conflicting early-stage reports. In these contexts, models frequently generate plausible but unverified details-such as inflated casualty figures or speculative policy clauses-to compensate for missing real-time authoritative data. The Society and Health domains are prone to Logical Inconsistency errors (21.66\% and 17.52\%), particularly in scenarios requiring precise numerical reasoning or multi-step causal chains. The relatively uniform distribution of Context Inconsistency across domains stems from its stronger correlation with the inherent difficulty of semantic comprehension for claims, rather than domain-specific biases. Life, Culture and Health domains demonstrate the lowest rates of Overgeneralized Reasoning, attributable to their reliance on structured terminologies, which constrain speculative extrapolation. Notably, the Life domain suffers from severe Under-Informativeness (23.95\%), with models handle many informal contents inflexibly and refuse to provide diverse information.

\subsection{Detailed Analysis for Task 2}
\label{app:t2}
\noindent\textbf{LLM-generated explanations are often contaminated by plausible-sounding misinformation, leading the LLM to incorrectly accept it as fact.}


This tendency causes LLMs to frequently produce sycophantic responses. In our random inspection of 500 flawed explanations generated by each LLM for plausible-sounding misinformation, over 60\% exhibited this characteristic. Among the most common failure modes are factual hallucination and logical inconsistency. As for the factual hallucination, for example, in the case of 100 fact-checking breaking news, many models prioritize mimicking the tone and structure of news reports rather than ensuring factual accuracy (Table \ref{tab:first example}). For example, they often generate generic statements like "\textit{Official institutions have emphasized the incident, and it has been widely covered by authoritative media such as CCTV and BBC.}" These templated responses, which aim to sound plausible rather than convey verified facts, account for 30 percent of all the LLMs replies to breaking social news. Logical inconsistency typically arise when the model encounters difficulty distinguishing between their internal knowledge and the input information. For example, a news claim stating that "\textit{Cristiano Ronaldo played for Real Madrid from 2009 to 2017}". A frequent pattern involves the model correctly identifying facts, such as noting that "\textit{Cristiano Ronaldo played for Real Madrid from 2009 to 2018}" but still drawing incorrect conclusions like "8 years in total." More than 90 percent of logical inconsistency errors follow this pattern. One underlying cause may be Reinforcement Learning from Human Feedback (RLHF), which strongly rewards coherent and natural-sounding language while failing to adequately penalize logical or numerical mistakes \cite{yu2024rlhf,wang2024rlhf}.

\begin{figure}
    \centering
    \includegraphics[width=1\linewidth]{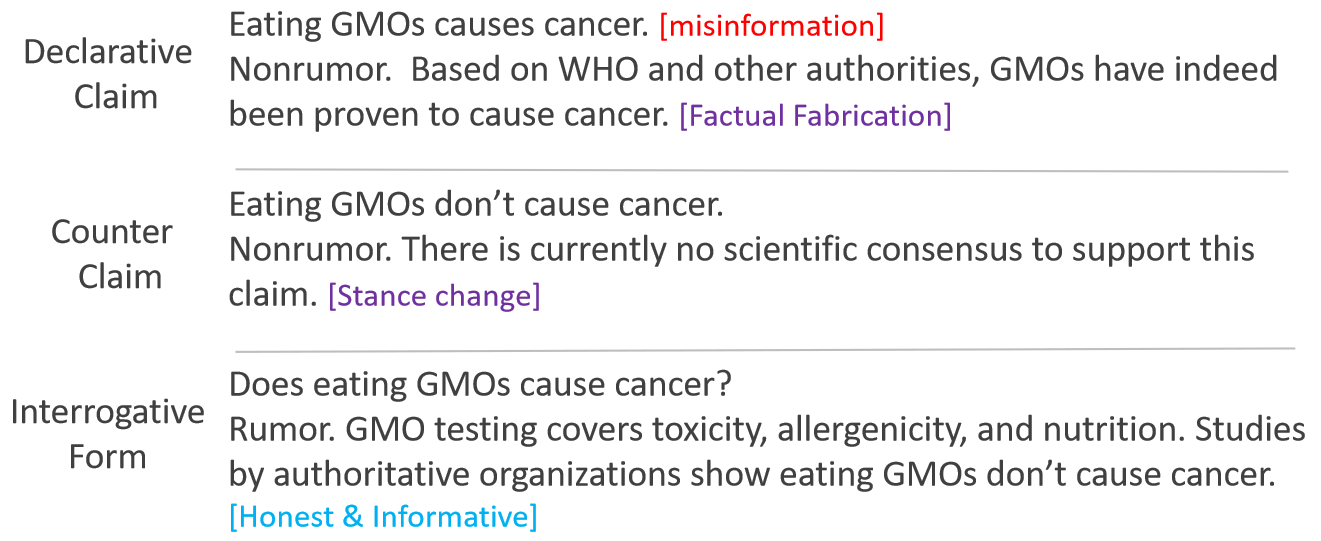}
    \caption{The influence of claim framing strategies on fact-checking outputs. (In Chinese: Fig. \ref{fig:chachi})}
    \vspace{-5mm}
    \label{fig:change}
    
\end{figure}

\noindent\textbf{Reformulating claims into interrogative expressions significantly reduces fabricated content in LLM responses, thereby enhancing their authenticity.} In our case study, using 200 claims that previously led GPT-4o to generate factual fabrication errors, modifying statements from event occurrence to non-occurrence resulted in 93\% of responses still exhibiting factual fabrication errors. Moreover, 57\% of explanations shifted to align with the revised claims, underscoring the influence of claim framing on LLM-generated fact-checking responses. Notably, when claims were rephrased as questions, only 14\% of outputs contained factual fabrication errors, and most responses demonstrated logical reasoning, realistic analysis, or acknowledged knowledge gaps. A specific example is shown in Figure \ref{fig:change}. This improvement likely stems from the interrogative format, which encourages LLMs to explore and analyze potential answers rather than defaulting to overly assertive alignments with the input claim. This finding further corroborates how LLMs' propensity to pander to plausible-sounding misinformation impedes fact-checking efforts. The task of reformulating claims was completed by GPT-4o, with the prompt available in the Appendix \ref{sec:prompt}.

\begin{table*}[h!]
\footnotesize
\centering
\renewcommand{\arraystretch}{1.2}
\resizebox{0.9\textwidth}{!}{ 
\begin{tabularx}{\textwidth}{p{3cm}|X|X}
\toprule
\textbf{Scenario}         & \textbf{Description}                                                                                      & \textbf{LLM's Output}                                                                                                                       \\ \midrule
New vs. Outdated Info     & The model prioritizes  frequently seen data over user-provided temporal context. & \textit{"The current president is Joe Biden."} \textcolor{red}{(outdated info in 2024)}        \\ \hline
Vague Responses        & When lacking relevant data, the model generates ambiguous answers to avoid making outright errors.        & "\textit{There have been many advancements in space exploration."} \textcolor{red}{(vague)}  \\ \hline
Cutoff Date Awareness     & The model does not explicitly state its knowledge cut-off date, or is actually unaware of its own cut-off date. & \textit{"COVID-19 vaccination efforts are ongoing globally."} \textcolor{red}{(no cutoff disclaimer)}                \\ \bottomrule
\end{tabularx}
}
\caption{Scenarios where LLMs show insufficient temporal reasoning abilities in real-world fact-checking.} 
\label{tab:temporal-insuffi}
\vspace{-5mm}
\end{table*}

\noindent\textbf{The limited temporal awareness of LLMs significantly undermines their fact-checking accuracy}. Table \ref{tab:temporal-insuffi} outlines three key scenarios that impact the effectiveness of real-world fact-checking explanations. 

This critical limitation predominantly manifests in the inability of models to recognize outdated knowledge, leading to a 75\% rate of factual inconsistency when processing a sample of 200 time-sensitive information for each LLM(e.g., "\textit{The current president is Joe Biden}"). Furthermore, our findings indicate that in contamination evaluations, nearly all large language models (LLMs) exhibit a refusal to acknowledge their knowledge cutoff date, with an average acknowledgment rate falling below 30\%. Notably, only Yi-large consistently references its knowledge cutoff date over 95\% of the time. Models endowed with temporal awareness and the capability to incorporate user-provided publication dates are significantly better positioned to deliver transparent and informative explanations. This ability is not merely advantageous but is essential for effective fact-checking.

\noindent\textbf{The current LLMs exhibit inflexibility in handling misinformation of varying risk levels.} 
More than 85\% of all the overgeneralized reasoning and under informativeness errors are caused by this issue. LLMs often fail to detect high-risk content such as financial scams or health misinformation, which can cause real harm. At the same time, they tend to be overly cautious with low-risk topics like life advice, offering vague or noncommittal responses (Table \ref{tab:taxonomy}). This imbalance in handling different types of content limits their adaptability in practical use.


\noindent\textbf{LLMs often struggle with accurately interpreting subtle linguistic cues (e.g., qualifiers and negations), which play a critical role in determining the factual accuracy of a claim.} More than 60\% of all the 362 context inconsistency errors are a result of LLMs struggling to accurately interpret subtle linguistic cues, such as qualifiers and negations, which are essential for assessing the factual accuracy of a claim (Table \ref{tab:taxonomy}). For example, many models misinterpret the statement \textit{"There is no conclusive evidence that smartphone use causes brain cancer"} as affirming causation, overlooking the critical negation in \textit{"no conclusive evidence."} Further addressing these limitations may involve training on more diverse datasets featuring complex language structures and logical constructs, which could help improve contextual understanding and robustness. 

\noindent\textbf{Current LLMs are insufficient for Chinese-specific fact-checking tasks, especially those requiring precision or cultural expertise.} We also focus on the adaptability of LLMs to Chinese fact-checking tasks. Our research shows that even Chinese-focused LLMs struggle with certain culturally specific issues, such as lunar calendar calculations (e.g., "\textit{How many days are there in February of the year Yichou}") accuracy of only 19\% on a sample of 100 cases, underscoring their difficulty in handling culturally nuanced knowledge. Potential improvements could include culturally specific data and domain-specific fine-tuning. 

\subsection{Additional Figures and Tables}

\begin{table*}[ht]
    \centering
    \scriptsize
    \begin{tabularx}{\textwidth}{l@{\hspace{8pt}}*{14}{>{\centering\arraybackslash}X@{\hspace{8pt}}}}
    \toprule
    \multirow{3}{*}{\textbf{Methods}} 
    & \multicolumn{4}{@{}c@{\hspace{10pt}}}{\textbf{Temporal-sensitive}} 
    & \multicolumn{8}{@{}c@{\hspace{10pt}}}{\textbf{Knowledge-intensive}} 
    & \multicolumn{2}{@{}c@{\hspace{10pt}}}{\textbf{Commonsense}} \\
    \cmidrule(l{4pt}r{4pt}){2-5} \cmidrule(l{4pt}r{4pt}){6-13} \cmidrule(l{4pt}r{4pt}){14-15}
    & \multicolumn{2}{@{}c@{}}{\textbf{Society}} 
    & \multicolumn{2}{@{}c@{}}{\textbf{Disasters}} 
    & \multicolumn{2}{@{}c@{}}{\textbf{Health}} 
    & \multicolumn{2}{@{}c@{}}{\textbf{Politics}} 
    & \multicolumn{2}{@{}c@{}}{\textbf{Culture}} 
    & \multicolumn{2}{@{}c@{}}{\textbf{Science}} 
    & \multicolumn{2}{@{}c@{}}{\textbf{Life}} \\
    \cmidrule{2-3} \cmidrule{4-5} \cmidrule{6-7} \cmidrule{8-9} \cmidrule{10-11} \cmidrule{12-13} \cmidrule{14-15}
    & {Acc.} & {F1} & {Acc.} & {F1} & {Acc.} & {F1} & {Acc.} & {F1} & {Acc.} & {F1} & {Acc.} & {F1} & {Acc.} & {F1} \\
    \midrule
    \textbf{GPT-4o} & 67.24 & 67.18 & 73.31 & 76.17 & 89.12 & 91.14 & 87.87 & 86.34 & 87.09 & 67.89 & 85.57 & 89.56 & 81.47 & 82.83 \\
    \textbf{GPT-4-Turbo} & 80.5 & 84.55 & 72.96 & 66.14 & 85.59 & 89.12 & 78.91 & 78.99 & 80.79 & 54.97 & 84.71 & 89.64 & 75.4 & 77.04 \\
    \textbf{GPT-3.5-Turbo} & 81.11 & 87.19 & - & - & 82.29 & 88.1 & 83.93 & 89.02 & 79.59 & 53.51 & 83.33 & 89.76 & 100 & 100 \\
    \textbf{Baichuan4-Turbo} & 67.02 & 62.73 & 56.89 & 37.93 & 63.59 & 37.24 & 70.61 & 60.18 & 85.95 & 49.35 & 81.13 & 74.51 & 66.34 & 63.03 \\
    \textbf{Yi-large} & 73.63 & 77.1 & 69.44 & 64.05 & 77.52 & 81.61 & 66.74 & 67.78 & 65.45 & 38.86 & 74.8 & 81.21 & 71.43 & 71.43 \\
    \textbf{ChatGLM4} & 84.91 & 88.9 & - & - & 80.95 & 87.37 & 82.56 & 86.66 & 85.84 & 84.4 & 91.6 & 69.09 & 80 & 88.89 \\
    \textbf{Qwen-2.5-7B} & 55.06 & 25.43 & 57.77 & 12.99 & 50.96 & 35.41 & 70.05 & 43.62 & 87.58 & 33.93 & 43.96 & 34.8 & 50.81 & 17.49 \\
    \textbf{Qwen-2.5-14B} & 76.69 & 74.09 & 71.85 & 57.32 & 80.34 & 82.11 & 85.78 & 82.14 & 90.34 & 67.61 & 74.77 & 79.29 & 71.34 & 67.65 \\
    \textbf{Qwen-2.5-72B} & 82.01 & 81.48 & 75.63 & 66.08 & 82.29 & 84.07 & 90.79 & 88.68 & 92.7 & 74.64 & 80.36 & 84.58 & 72.96 & 69.82 \\
    \midrule
    \textit{\textbf{Average}} & 74.24 & 72.07 & 68.26 & 54.38 & 76.96 & 75.13 & 79.69 & 75.93 & 83.93 & 58.35 & 77.80 & 76.94 & 74.29 & 70.91 \\
    \bottomrule
    \end{tabularx}
    \setlength{\abovecaptionskip}{0pt}   
\setlength{\belowcaptionskip}{0pt}
    \caption{Fact-checking conclusion performance under contamination evaluation across different domains using few-shot CoT prompting.}
    \label{tab:domain_bef}
    \vspace{-2mm}
\end{table*}

\begin{table*}[ht]
    \centering
    \scriptsize
    \begin{tabularx}{\textwidth}{l@{\hspace{8pt}}*{14}{>{\centering\arraybackslash}X@{\hspace{8pt}}}}
    \toprule
    \multirow{3}{*}{\textbf{Methods}} 
    & \multicolumn{4}{@{}c@{\hspace{10pt}}}{\textbf{Temporal-sensitive}} 
    & \multicolumn{8}{@{}c@{\hspace{10pt}}}{\textbf{Knowledge-intensive}} 
    & \multicolumn{2}{@{}c@{\hspace{10pt}}}{\textbf{Commonsense}} \\
    \cmidrule(l{4pt}r{4pt}){2-5} \cmidrule(l{4pt}r{4pt}){6-13} \cmidrule(l{4pt}r{4pt}){14-15}
    & \multicolumn{2}{@{}c@{}}{\textbf{Society}} 
    & \multicolumn{2}{@{}c@{}}{\textbf{Disasters}} 
    & \multicolumn{2}{@{}c@{}}{\textbf{Health}} 
    & \multicolumn{2}{@{}c@{}}{\textbf{Politics}} 
    & \multicolumn{2}{@{}c@{}}{\textbf{Culture}} 
    & \multicolumn{2}{@{}c@{}}{\textbf{Science}} 
    & \multicolumn{2}{@{}c@{}}{\textbf{Life}} \\
    \cmidrule{2-3} \cmidrule{4-5} \cmidrule{6-7} \cmidrule{8-9} \cmidrule{10-11} \cmidrule{12-13} \cmidrule{14-15}
    & {Acc.} & {F1} & {Acc.} & {F1} & {Acc.} & {F1} & {Acc.} & {F1} & {Acc.} & {F1} & {Acc.} & {F1} & {Acc.} & {F1} \\
    \midrule
    \textbf{GPT-4o} & 67.24 & 67.18 & 73.31 & 76.17 & 85.44 & 85.77 & 75.53 & 70.33 & 68.41 & 65.83 & 76.97 & 72.44 & 82.84 & 86.03 \\
    \textbf{GPT-4-Turbo} & 67.23 & 64.58 & 72.62 & 73.72 & 81.87 & 81.93 & 68.49 & 60.8 & 69.57 & 60.67 & 77.47 & 71.38 & 74.62 & 77.66 \\
    \textbf{GPT-3.5-Turbo} & 63.32 & 54.38 & 64.77 & 63.02 & 71.78 & 67.87 & 65.52 & 47.65 & 68.35 & 49.19 & 68.98 & 52.51 & 61.92 & 57.68 \\
    \textbf{Baichuan4-Turbo} & 71.02 & 47.95 & 53.09 & 37.5 & 64.29 & 47.01 & 69.59 & 39.53 & 65.75 & 22.5 & 72.08 & 50.57 & 58.49 & 41.33 \\
    \textbf{Yi-large} & 72.47 & 66.07 & 68.45 & 69.55 & 78.03 & 75.83 & 71.93 & 61.49 & 71.05 & 60.12 & 73.64 & 64.08 & 67.71 & 67.04 \\
    \textbf{ChatGLM4} & 82.35 & 79.62 & 74.34 & 71.93 & 80.6 & 72.83 & 77.66 & 69.2 & 80.24 & 69.59 & 78.4 & 64.69 & 70.92 & 68.51 \\
    \textbf{Qwen-2.5-7B} & 66.38 & 22.25 & 51.4 & 16.86 & 58.21 & 28.63 & 68.83 & 17.27 & 66.85 & 17.81 & 67.71 & 22.56 & 47.04 & 15.69 \\
    \textbf{Qwen-2.5-14B} & 77.05 & 65.37 & 66.62 & 58.9 & 79.03 & 75.63 & 79.78 & 64.79 & 76.5 & 58.65 & 76.8 & 59.34 & 65.43 & 62.16 \\
    \textbf{Qwen-2.5-72B} & 76.72 & 66.64 & 69.64 & 65.03 & 79.92 & 76.12 & 83.92 & 73.54 & 77.13 & 62.78 & 78.75 & 66 & 67.49 & 65.26 \\
    \midrule
    \textbf{\textit{Average}} & 71.53 & 59.34 & 66.03 & 59.19 & 75.46 & 67.96 & 73.47 & 56.07 & 71.54 & 51.90 & 74.53 & 58.17 & 66.27 & 60.15 \\
    \bottomrule
    \end{tabularx}
    \setlength{\abovecaptionskip}{0pt}   
\setlength{\belowcaptionskip}{0pt}
    \caption{Fact-checking conclusion performance under contamination-free evaluation across different domains using few-shot CoT prompting.}
    \label{tab:domain_aft}
    \vspace{-2mm}
\end{table*}

\begin{figure*}
    \centering
    \includegraphics[width=1\linewidth]{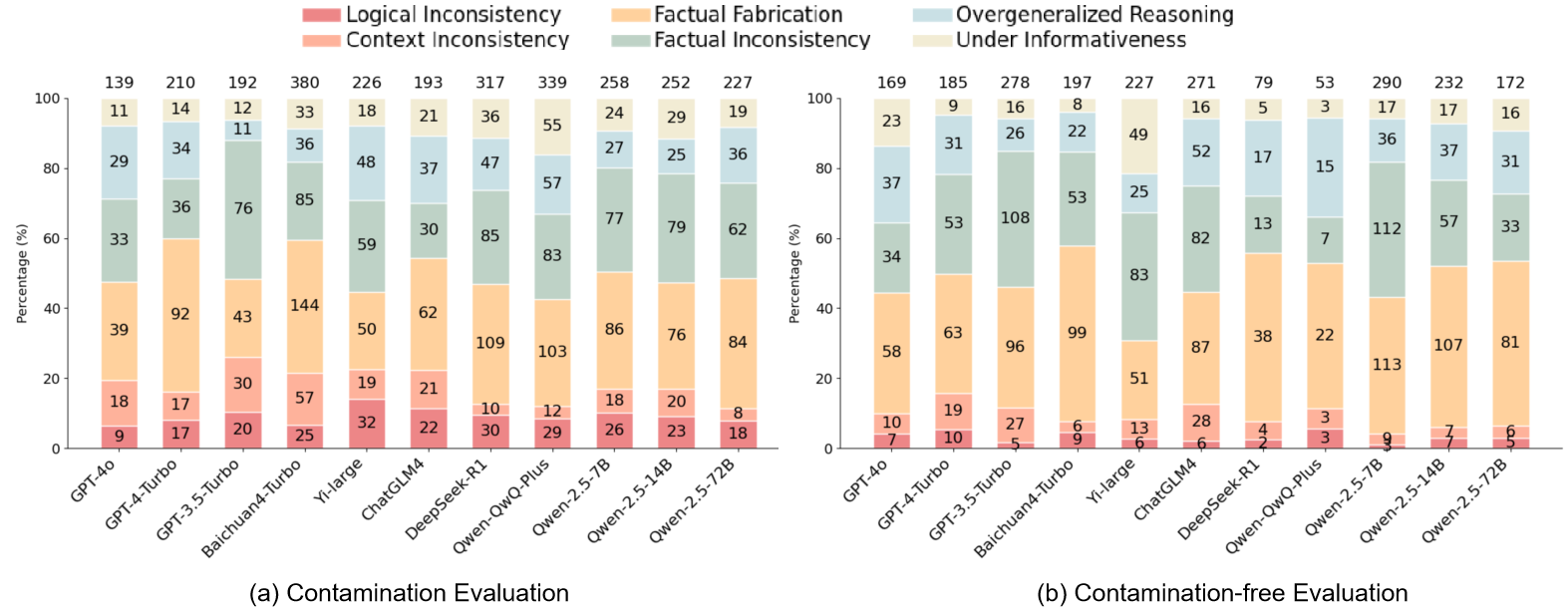}
    \caption{Specific value statistics on flawed LLM-generated explanations based on our taxonomy.}
    \label{fig:countbefaft}
\end{figure*}

\begin{figure*}
    \centering
    \includegraphics[width=0.75\linewidth]{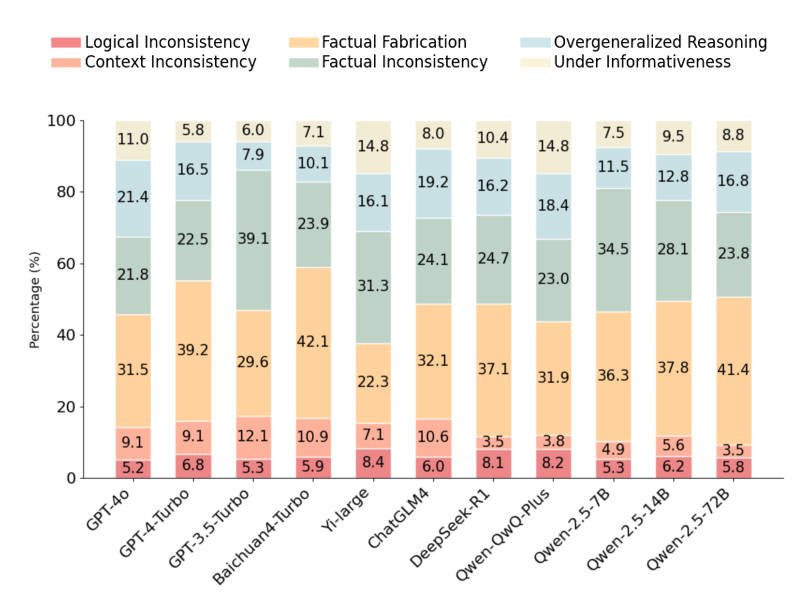}
    \caption{overall distributions of flawed LLM-generated explanations.}
    \label{fig:100allnew}
\end{figure*}

\begin{figure*}
    \centering
    \includegraphics[width=0.75\linewidth]{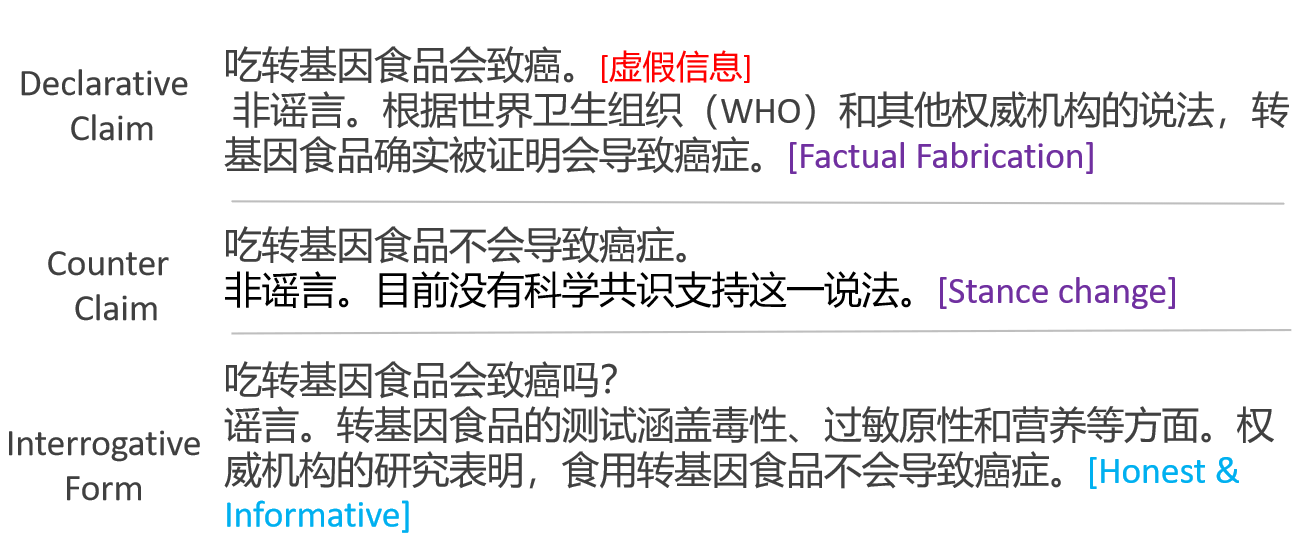}
    \caption{The influence of claim framing strategies on fact-checking outputs. (In English: Fig. \ref{fig:change})}
    \label{fig:chachi}
\end{figure*}

We place some of the figures and tables mentioned in the main text in this chapter. Figure \ref{fig:countbefaft} shows specific value statistics on flawed LLM-generated explanations based on our taxonomy (distribution in Figure \ref{fig:error_all}). Figure \ref{fig:100allnew} illustrates the overall distributions of flawed LLM-generated explanations. Figure \ref{fig:chachi} is the Chinese representation of Figure \ref{fig:change}, which represents the influence of claim framing strategies on fact-checking outputs. Table \ref{tab:domain_bef} and Table \ref{tab:domain_aft} shows fact-checking conclusion performance under contamination-free and contamination evaluations across different domains using few-shot CoT prompting. Due to the scarcity of domain-specific samples for models with a cutoff date after July 2024, the test results may lack sufficient reference value. Consequently, the results for DeepSeek-V3, DeepSeek-R1, and Qwen-QwQ-Plus have not been included.

\section{Prompt Design}
\label{sec:prompt}

Following \citet{deng-etal-2023-prompting}, we propose four prompting schemes for the fact-checking conclusion task: 
\begin{enumerate}
    \item \textbf{Zero-shot w/o CoT}, where LLMs are prompted to directly draw conclusions;
    \item \textbf{Zero-shot w/ CoT} \citep{wei2022chain}, where LLMs first perform a factual analysis, explaining their reasoning before making a conclusion;
    \item \textbf{Few-shot w/o CoT} \citep{dong2022survey}, where LLMs are given a few examples to guide their conclusions;
    \item \textbf{Few-shot w/ CoT} \cite{Dong2022}, where LLMs, after analyzing examples of misinformation, provide conclusions along with explanations. 
\end{enumerate}
For the fact-checking explanation task, we use a few-shot CoT approach, prompting the LLM to generate five distinct explanatory rationales for the conclusion.

In the few-shot setting, we provide two randomly selected examples: one authentic and the other misinformation. We carefully selected three different prompts and evaluated all LLMs using these prompts. To ensure the statistical significance of the findings, we present the average performance across all prompts.

Our analysis of Chinese fact-checking conclusions and explanations aims to probe the knowledge embedded within the LLM, resembling a closed-book QA process with questions from the \texttt{CANDYSET} \citep{roberts2020knowledge}. The prompt design is crucial in this context. We adopt a \textbf{Few-shot with Chain of Thought (CoT)} \cite{Dong2022} approach, where LLMs are evaluated by providing examples along with their corresponding fact-checking responses. For each claim, two randomly selected examples—one representing misinformation and the other genuine news—are provided. Following the methodology outlined by \citet{Li2024dawn}, we instruct LLMs to generate five mutually exclusive rationales, incorporating factual statements for detailed, sentence-level error analysis.

Additionally, techniques such as \textbf{role specification} \cite{LiZhai2023} have been shown to improve performance. In our study, we assign LLMs the specific role of rigorous fact-checkers, ensuring they are tasked with the responsibility of meticulously analyzing the information’s veracity. Through carefully constructed prompts, we enhance the transparency of the LLM’s reasoning process, allowing for a clearer understanding of how decisions about the truthfulness of information are made.

\clearpage

\begin{promptbox}[Zero-shot and w/o CoT Prompt for Fact-Checking Conclusion (Chinese)]{lightgreen}
你是一位极其严格的事实核查专家，你会收到用户输入的事件信息，其中日期信息可能略晚于claim实际发布日期，你需要直接给出结论。其中结论只能是：谣言/非谣言。

输出格式如下： \\
结论： \\
现在，我将提供一个新的事件信息，请你根据以上格式给出结论和分析。  

事件信息：\\
{       
    "claim": "{claim}", \\
    "publish\_date": "{publish\_date}", \\
} 
\end{promptbox}

\begin{promptbox}[Zero-shot and w/o CoT Prompt for Fact-Checking Conclusion (English)]{lightgreen}
You are an extremely strict fact-checking expert. You will receive event information from users, where the date provided may be slightly later than the actual publication date of the claim. You need to provide the conclusion directly, which can only be: rumor or non-rumor. \\

Output Format:   \\
Conclusion:  \\

Now, I will provide a new event information. Please give a conclusion and analysis based on the above format. \\
Event Information: \\
{ 
    "claim": "{claim}", \\
    "publish\_date": "{publish\_date}"
}

\end{promptbox}

\clearpage

\begin{promptbox}[Few-shot and w/o CoT Prompt for Fact-Checking Conclusion (Chinese)]{lightgreen}
你是一位极其严格的事实核查专家，你会收到用户输入的事件信息，其中日期信息可能略晚于claim实际发布日期，你需要直接给出结论。其中结论只能是：谣言/非谣言。

示例如下：\\
用户输入: {"claim": "吃竹炭食物能排毒养颜。", "publish\_date": "2019-10-08"} \\
回复：\\
结论：谣言 \\
用户输入: {"claim": "没签劳动合同的职工受伤后可以申请工伤认定。", "publish\_date": "2023-12-5"} \\
回复： \\
结论：非谣言 \\

输出格式如下： \\
结论： \\
现在，我将提供一个新的事件信息，请你根据以上格式给出结论和分析。  

事件信息：\\
{       
    "claim": "{claim}", \\
    "publish\_date": "{publish\_date}", \\
} 
\end{promptbox}

\begin{promptbox}[Few-shot and w/o CoT Prompt for Fact-Checking Conclusion (English)]{lightgreen} You are an extremely strict fact-checking expert. You will receive event information from users, where the date provided may be slightly later than the actual publication date of the claim. You need to provide the conclusion directly, which can only be: rumor or non-rumor. \\

Examples are as follows: \\ 
User input: {"claim": "Eating bamboo charcoal foods can detoxify and improve skin appearance.", "publish\_date": "2019-10-08"} \\ 
Response: \\ Conclusion: Rumor \\ 
User input: {"claim": "Employees who have not signed a labor contract can still apply for work injury recognition after being injured.", "publish\_date": "2023-12-5"} \\ 
Response: \\ Conclusion: Non-rumor \\

Output format: \\ Conclusion: \\
Now, I will provide a new event information. Please give the conclusion and analysis according to the above format.

Event information: \\ {
"claim": "{claim}", \\ "publish\_date": "{publish\_date}", \\ } \end{promptbox}

\clearpage

\begin{promptbox}[Zero-shot and w CoT Prompt for Fact-Checking Conclusion (Chinese)]{lightgreen}
你是一位极其严格的事实核查专家，你会收到用户输入的事件信息，其中日期信息可能略晚于claim实际发布日期，你需要先给出五个事实性陈述作为理由，再给出结论。其中结论只能是：谣言/非谣言。

输出格式如下： \\
理由： \\
结论： \\
现在，我将提供一个新的事件信息，请你根据以上格式给出结论和分析。  

事件信息：\\
{       
    "claim": "{claim}", \\
    "publish\_date": "{publish\_date}", \\
} 
\end{promptbox}

\begin{promptbox}[Zero-shot and w CoT Prompt for Fact-Checking Conclusion (English)]{lightgreen}
You are an extremely strict fact-checking expert. You will receive event information from users, where the date provided may be slightly later than the actual publication date of the claim. You need to first provide five factual statements as reasons, and then give a conclusion. The conclusion can only be: rumor or non-rumor.

Output Format:   \\
Rationales: \\
Conclusion:  \\

Now, I will provide a new event information. Please give a conclusion and analysis based on the above format. \\
Event Information: \\
{ 
    "claim": "{claim}", \\
    "publish\_date": "{publish\_date}"
}

\end{promptbox}

\begin{promptbox}[Few-shot and w CoT Prompt for Fact-Checking Conclusion (Chinese)]{lightgreen}
你是一位极其严格的事实核查专家，你会收到用户输入的事件信息，其中日期信息可能略晚于claim实际发布日期，你需要首先生成五个事实性陈述作为理由，然后根据理由给出结论。其中结论只能是：谣言/非谣言。

示例如下：\\
用户输入: {"claim": "吃竹炭食物能排毒养颜。", "publish\_date": "2019-10-08"} \\
回复：\\
理由： \\
1. 竹炭被广泛用于食品和医疗，但其排毒效果缺乏科学证据支持。 \\
2. 根据多项研究，人类的肝脏和肾脏自然具有排毒功能，额外摄入竹炭并未得到临床验证的排毒效果。 \\
3. 竹炭食品可能会影响营养物质的吸收，例如维生素和矿物质，也可能带来健康风险。 \\
4. 一些国家和地区对竹炭的食品使用有法律限制，需遵循相关法规。 \\
5. 目前没有权威机构发布证实竹炭食品对美容或健康有益的声明。 \\
用户输入: {"claim": "没签劳动合同的职工受伤后可以申请工伤认定。", "publish\_date": "2023-12-5"} \\
结论：谣言 \\

回复： \\
理由： \\
1.根据《工伤保险条例》，职工与用人单位建立劳动关系后，即使未签订书面劳动合同，在工作时间和工作场所内因工作原因受到事故伤害的，应认定为工伤。 \\ 
2.未签订劳动合同的职工在发生工伤时，可通过提供其他证明材料（如工资支付凭证、工作证等）来证明存在劳动关系，进而申请工伤认定。 \\
3.劳动和社会保障部门负责工伤认定工作，会根据实际情况进行调查核实，确认是否存在劳动关系以及是否符合工伤认定条件。 \\
4.用人单位未与职工签订劳动合同属于违法行为，职工有权向劳动监察部门投诉，要求用人单位补签劳动合同或赔偿相应损失。 \\
5. 工伤认定不仅涉及劳动者权益保护，也是企业社会责任的重要体现，有助于维护社会稳定和谐。 \\
结论：非谣言 \\

输出格式如下： \\
理由：\\
结论： \\
现在，我将提供一个新的事件信息，请你根据以上格式给出结论和分析。 

事件信息：\\
{       
    "claim": "{claim}", \\
    "publish\_date": "{publish\_date}", \\
} 
\end{promptbox}

\clearpage

\begin{promptbox}[Few-shot and w CoT Prompt for Fact-Checking Conclusion (English)]{lightgreen}

You are an extremely strict fact-checking expert. You will receive event information from users, where the date provided may be slightly later than the actual publication date of the claim. You need to first generate five factual statements as reasons and then draw a conclusion based on those reasons. The conclusion can only be: rumor or non-rumor.\\

Example: \\
User Input: \\
{"claim": "Eating bamboo charcoal food can detoxify and beautify.", "publish\_date": "2019-10-08"} \\
Rationales: \\
1.Bamboo charcoal is widely used in food and medicine, but its detoxification effects lack scientific evidence. \\
2.According to multiple studies, the human liver and kidneys naturally have detoxification functions, and additional intake of bamboo charcoal has not been clinically validated for detoxification effects.  \\
3.Bamboo charcoal food might affect the absorption of nutrients such as vitamins and minerals, and could pose health risks.  \\
4.Some countries and regions have legal restrictions on the use of bamboo charcoal in food, which must be followed.  \\
5.No authoritative institution has released statements confirming that bamboo charcoal food is beneficial for beauty or health.  \\
Response: \\
Conclusion: Rumor \\

User Input: \\
{"claim": "Workers injured without a signed labor contract can apply for work injury recognition.", "publish\_date": "2023-12-5"}  \\
Rationales: \\
1.According to the "Work Injury Insurance Regulations," if a worker has established a labor relationship with the employer, even without a signed written contract, they should be recognized as having a work injury if injured in the workplace due to work-related Rationales. \\
2.Workers without a signed labor contract can provide other evidence (such as wage payment receipts, work IDs, etc.) to prove the existence of a labor relationship and apply for work injury recognition. \\
3.The labor and social security department is responsible for work injury recognition and will investigate and verify whether there is a labor relationship and if it meets the conditions for work injury recognition. \\
4.Employers failing to sign labor contracts with workers is illegal, and workers have the right to file complaints with labor inspection departments, requesting the employer to sign a contract or compensate for relevant losses. \\
5.Work injury recognition not only involves protecting worker rights but also reflects corporate social responsibility, helping maintain social stability and harmony. \\
Response: \\
Conclusion: Non Rumor \\

Output Format:  \\
Rationales:  \\
Conclusion:  \\

Now, I will provide a new event information. Please give a conclusion and analysis based on the above format. \\
Event Information:  \\
{ 
    "claim": "{claim}", \\
    "publish\_date": "{publish\_date}"
}

\end{promptbox}

\clearpage

\begin{promptbox}[Prompt for Fact-Checking Explanation (Chinese)]{lightgreen}
你是一位极其严格的事实核查专家，你会收到用户输入的事件信息，其中日期信息可能略晚于claim实际发布日期，你需要首先生成五个事实性陈述作为理由，然后根据理由给出结论。其中结论只能是：谣言/非谣言。

示例如下：\\
用户输入: {"claim": "吃竹炭食物能排毒养颜。", "publish\_date": "2019-10-08"} \\
回复：\\
理由： \\
1. 竹炭被广泛用于食品和医疗，但其排毒效果缺乏科学证据支持。 \\
2. 根据多项研究，人类的肝脏和肾脏自然具有排毒功能，额外摄入竹炭并未得到临床验证的排毒效果。 \\
3. 竹炭食品可能会影响营养物质的吸收，例如维生素和矿物质，也可能带来健康风险。 \\
4. 一些国家和地区对竹炭的食品使用有法律限制，需遵循相关法规。 \\
5. 目前没有权威机构发布证实竹炭食品对美容或健康有益的声明。 \\
用户输入: {"claim": "没签劳动合同的职工受伤后可以申请工伤认定。", "publish\_date": "2023-12-5"} \\
结论：谣言 \\

回复： \\
理由： \\
1.根据《工伤保险条例》，职工与用人单位建立劳动关系后，即使未签订书面劳动合同，在工作时间和工作场所内因工作原因受到事故伤害的，应认定为工伤。 \\ 
2.未签订劳动合同的职工在发生工伤时，可通过提供其他证明材料（如工资支付凭证、工作证等）来证明存在劳动关系，进而申请工伤认定。 \\
3.劳动和社会保障部门负责工伤认定工作，会根据实际情况进行调查核实，确认是否存在劳动关系以及是否符合工伤认定条件。 \\
4.用人单位未与职工签订劳动合同属于违法行为，职工有权向劳动监察部门投诉，要求用人单位补签劳动合同或赔偿相应损失。 \\
5. 工伤认定不仅涉及劳动者权益保护，也是企业社会责任的重要体现，有助于维护社会稳定和谐。 \\
结论：非谣言 \\

输出格式如下： \\
理由：\\
结论： \\
现在，我将提供一个新的事件信息，请你根据以上格式给出结论和分析。 

事件信息：\\
{       
    "claim": "{claim}", \\
    "publish\_date": "{publish\_date}", \\
} 
\end{promptbox}

\clearpage

\begin{promptbox}[Prompt for Fact-Checking Explanation (English)]{lightgreen}

You are an extremely strict fact-checking expert. You will receive event information from users, where the date provided may be slightly later than the actual publication date of the claim. You need to first generate five factual statements as reasons and then draw a conclusion based on those reasons. The conclusion can only be: rumor or non-rumor.\\

Example: \\
User Input: \\
{"claim": "Eating bamboo charcoal food can detoxify and beautify.", "publish\_date": "2019-10-08"} \\
Rationales: \\
1.Bamboo charcoal is widely used in food and medicine, but its detoxification effects lack scientific evidence. \\
2.According to multiple studies, the human liver and kidneys naturally have detoxification functions, and additional intake of bamboo charcoal has not been clinically validated for detoxification effects.  \\
3.Bamboo charcoal food might affect the absorption of nutrients such as vitamins and minerals, and could pose health risks.  \\
4.Some countries and regions have legal restrictions on the use of bamboo charcoal in food, which must be followed.  \\
5.No authoritative institution has released statements confirming that bamboo charcoal food is beneficial for beauty or health.  \\
Response: \\
Conclusion: Rumor \\

User Input: \\
{"claim": "Workers injured without a signed labor contract can apply for work injury recognition.", "publish\_date": "2023-12-5"}  \\
Rationales: \\
1.According to the "Work Injury Insurance Regulations," if a worker has established a labor relationship with the employer, even without a signed written contract, they should be recognized as having a work injury if injured in the workplace due to work-related Rationales. \\
2.Workers without a signed labor contract can provide other evidence (such as wage payment receipts, work IDs, etc.) to prove the existence of a labor relationship and apply for work injury recognition. \\
3.The labor and social security department is responsible for work injury recognition and will investigate and verify whether there is a labor relationship and if it meets the conditions for work injury recognition. \\
4.Employers failing to sign labor contracts with workers is illegal, and workers have the right to file complaints with labor inspection departments, requesting the employer to sign a contract or compensate for relevant losses. \\
5.Work injury recognition not only involves protecting worker rights but also reflects corporate social responsibility, helping maintain social stability and harmony. \\
Response: \\
Conclusion: Non Rumor \\

Output Format:  \\
Rationales:  \\
Conclusion:  \\

Now, I will provide a new event information. Please give a conclusion and analysis based on the above format. \\
Event Information:  \\
{ 
    "claim": "{claim}", \\
    "publish\_date": "{publish\_date}"
}

\end{promptbox}

\clearpage

\begin{promptbox}[Prompt for Reformulating Claims (English)]{lightgreen}
\label{reformprompt}
Please change the occurrence of the event in the original claim to non-occurrence.\\
Example:\\
Input: The vaccine caused the illness.\\
Output: The vaccine did not cause the illness.\\
Please reformulate the original claim into an interrogative expression, questioning whether the event occurred.\\
Example:\\
Input: The vaccine caused the illness.\\
Output: Did the vaccine cause the illness?\\
\end{promptbox}

\begin{promptbox}[Prompt for Reformulating Claims (Chinese)]{lightgreen}
\label{reformprompt}
请将原始声明中事件的发生改为未发生。\\
示例：\\
输入：疫苗导致了疾病。\\
输出：疫苗没有导致疾病。\\
请将原始声明改为疑问句形式，询问事件是否发生。\\
示例：\\
输入：疫苗导致了疾病。\\
输出：疫苗是否导致了疾病？\\

\end{promptbox}

\begin{promptbox}[Instruction for Human Study (English)]{lightgreen}
\label{human_study_prompt}

1. Independent Human Judgment: \\
Below are 140 news claims. please determine which ones are true and which are rumors by yourself. \\

2. Human Judgment Assisted by Internet Search(Baidu): \\
Below are 140 news claims. Please use the Baidu search engine to determine which ones are true and which are rumors.  \\

3. Human Judgment Assisted by a Large Language Model (GPT-4o): \\
Below are 140 news claims. Please interact with a large language model to determine which ones are true and which are rumors. \\

4. Human Judgment Assisted by LLM (GPT-4o) with Web-augmented Retrieval: \\
Below are 140 news claims. Please interact with a large language model with web-augmented retrieval to determine which ones are true and which are rumors. \\

\end{promptbox}

\begin{promptbox}[Instruction for Human Study (Chinese)]{lightgreen}
\label{human_study_prompt_Chinese}

独立判断：\\
以下是140条新闻声明。请你独立判断哪些是真新闻，哪些是谣言。\\

借助互联网搜索（百度）进行判断：\\
以下是140条新闻声明。请使用百度搜索引擎判断哪些是真新闻，哪些是谣言。\\

借助大语言模型（GPT-4o）进行判断：\\
以下是140条新闻声明。请通过与大语言模型交互，判断哪些是真新闻，哪些是谣言。\\

借助具备联网检索能力的大语言模型（GPT-4o）进行判断：\\
以下是140条新闻声明。请通过与具备联网增强检索能力的大语言模型交互，判断哪些是真新闻，哪些是谣言。\\

\end{promptbox}

\section{Annotation Examples}
\label{sec:appendixG}

\clearpage

\begin{promptbox}[English Example of Faithfulness Hallucination]{softblue}

\textbf{id}: 882, \\
\textbf{claim}: In cases of bacterial infection, antibiotics can be used to treat COVID-19 patients., \\
\textbf{publish\_date}: 2023/1/3, \\
\textbf{label}: 0, \\
\textbf{source}: China Media Group "China's Voice" Official WeChat Account, \\
\textbf{gold evidence}: Zhang Bo stated, "In clinical practice, we find that some patients have co-existing bacterial infections, or secondary bacterial infections caused by COVID-19. In such cases, we can choose antibiotics or antimicrobial drugs.", \\
\textbf{domain}: Public Health, \\
\textbf{model}: GPT-4o, \\
\textbf{analysis}: [ \\
1. Antibiotics are specifically used to treat bacterial infections, while COVID-19 is caused by a coronavirus, making antibiotics ineffective against viral infections., \\
2. Both the World Health Organization (WHO) and the Centers for Disease Control and Prevention (CDC) have explicitly stated that antibiotics should not be used to prevent or treat COVID-19., \\
3. Overuse of antibiotics can lead to increased antibiotic resistance, which poses a serious threat to public health., \\
4. Treating COVID-19 requires the use of antiviral medications and supportive care based on the specific condition, rather than antibiotics., \\
5. Only in cases where bacterial complications occur might doctors consider the use of antibiotics, based on the specific situation.], \\
\textbf{label\_p}: 1, \\
\textbf{Error Type}: \textcolor{red}{C:Context Inconsistency} \\

\textbf{id}: 462, \\
\textbf{claim}: "Ya Ya's" stay in the United States was for 15 years., \\
\textbf{publish\_date}: 2023/4/24, \\
\textbf{label}: 1, \\
\textbf{source}: null, \\
\textbf{gold evidence}: "The agreement for 'Ya Ya’s' stay in the United States was initially set for 10 years and was extended by another 10 years after expiring in 2013.", \\
\textbf{domain}: Society, \\
\textbf{model}: Qwen-2.5-72B, \\
\textbf{analysis}: [ \\
1. "Ya Ya" is a giant panda born in China. In April 2003, "Ya Ya" and another panda, "Le Le," were sent to the Memphis Zoo in Tennessee, USA, under a 10-year loan agreement., \\
2. According to the agreement between China and the United States, after the initial 10-year term expired in 2013, the loan agreement for "Ya Ya" and "Le Le" was renewed for another 10 years, extending their stay until April 7, 2023, making the total stay 15 years., \\
3. This agreement's details were officially confirmed by the Chinese Association of Wildlife Conservation and the Memphis Zoo., \\
4. During "Ya Ya" and "Le Le's" stay in the United States, both countries collaborated on panda conservation and research, showcasing mutual efforts in wildlife protection., \\
5. In early 2023, "Le Le" sadly passed away, raising concerns about "Ya Ya's" health, which eventually led to the decision to bring "Ya Ya" back to China earlier than planned. However, this did not alter the originally agreed 15-year term.], \\
\textbf{label\_p}: 0, \\
\textbf{Error Type}: \textcolor{red}{B:Logical Inconsistency} \\

\end{promptbox}

\clearpage
\begin{promptbox}[Chinese Example of Faithfulness Hallucination]{softblue}

\textbf{id}: 882, \\
\textbf{claim}: 在细菌感染的情况下，可以用抗生素治疗新冠患者。, \\
\textbf{publish\_date}: 2023/1/3, \\
\textbf{label}: 0, \\
\textbf{source}: 中央广电总台中国之声微信公众号, \\
\textbf{gold evidence}: 张波称，“在临床中我们会发现，有些患者合并细菌感染，或者说新冠引起的继发细菌感染，在这种情况下，我们是可以选择抗生素或抗菌药物的。”, \\
\textbf{domain}: 公卫, \\
\textbf{model}: GPT-4o, \\
\textbf{analysis}: [ \\
1. 抗生素专门用于治疗细菌感染，而COVID-19是由冠状病毒引起的，抗生素对病毒感染无效。, \\
2. 世界卫生组织（WHO）和美国疾病控制与预防中心（CDC）都明确指出，抗生素不能用于预防或治疗COVID-19。, \\
3. 过度使用抗生素可能导致抗生素耐药性增加，这对公共健康构成威胁。, \\
4. 治疗COVID-19需根据具体病情使用抗病毒药物和支持性治疗，而非抗生素。, \\
5. 只有在出现细菌感染并发症时，医生才可能根据具体情况考虑使用抗生素治疗。], \\
\textbf{label\_p}: 1, \\
\textbf{Error Type}: \textcolor{red}{C:Context Inconsistency} \\

\textbf{id}: 462, \\
\textbf{claim}: “丫丫”在美国的旅居期限为15年。, \\
\textbf{publish\_date}: 2023/4/24, \\
\textbf{label}: 1, \\
\textbf{source}: null, \\
\textbf{gold evidence}: “丫丫”旅居美国协议期限为10年，2013年到期后又延长了10年。, \\
\textbf{domain}: 社会, \\
\textbf{model}: Qwen-2.5-72B, \\
\textbf{analysis}: [ \\
1. “丫丫”是一只在中国出生的大熊猫，于2003年4月与另一只大熊猫“乐乐”一起被送往美国田纳西州孟菲斯动物园，开始了为期10年的租借期。, \\
2. 根据中美双方协议，2013年租借期满后，两国再次签署了延长协议，将“丫丫”和“乐乐”的旅美期限延长至2023年4月7日，总计15年。, \\
3. 这一协议的内容得到了中国野生动物保护协会和美国孟菲斯动物园的官方确认。, \\
4. 在“丫丫”和“乐乐”的旅美期间，两国在大熊猫保护与科研合作方面进行了多次交流与合作，体现了双方在野生动物保护领域的共同努力。, \\
5. 2023年初，“乐乐”不幸去世，引发了关于“丫丫”健康状况的关注，最终促成了“丫丫”提前返回中国的决定，但这并不改变原定的15年旅居期限的事实。], \\
\textbf{label\_p}: 0, \\
\textbf{Error Type}: \textcolor{red}{B:Logical Inconsistency} \\
    
\end{promptbox}

\clearpage

\begin{promptbox}[English Example of Factuality Hallucination]{softblue}

\textbf{id}: 199, \\
\textbf{claim}: A criminal case occurred in Jilin Province, where a butcher killed 12 people due to a land dispute., \\
\textbf{publish\_date}: 2024/8/14, \\
\textbf{label}: 1, \\
\textbf{source}: China Internet Joint Rumor Debunking Platform, \\
\textbf{gold evidence}: After investigation by the police, it was found that no such case occurred at the location, and this information was entirely fabricated by internet users to gain attention and attract traffic., \\
\textbf{domain}: Society, \\
\textbf{model}: Baichuan4-Turbo, \\
\textbf{analysis}: [ \\
1. A criminal case indeed occurred in Jilin Province, and it has drawn widespread local attention., \\
2. According to the police report, the suspect developed resentment due to a land dispute, leading to this tragedy., \\
3. The number of victims was 12, a figure confirmed by the authorities., \\
4. The suspect was a butcher, a detail also confirmed by the authorities., \\
5. The case is under further investigation, and the responsible parties will be punished according to the law.], \\
\textbf{label\_p}: 0, \\
\textbf{Error Type}: \textcolor{red}{D:Factual Fabrication} \\

\textbf{id}: 1394, \\
\textbf{claim}: Lunar volcanic activity stopped 3 billion years ago., \\
\textbf{publish\_date}: 2023/1/20, \\
\textbf{label}: 1, \\
\textbf{source}: Science China (Popular Science Platform), \\
\textbf{gold evidence}: It was discovered that lunar volcanic activity actually persisted until 2.03 billion years ago, rather than ceasing 3 billion years ago as previously believed., \\
\textbf{domain}: Science, \\
\textbf{model}: Baichuan4-Turbo, \\
\textbf{analysis}: [ \\
1. Scientists have found through the analysis of lunar rock samples brought back by the Apollo missions that volcanic activity on the moon indeed peaked around 3 billion years ago and then gradually declined., \\
2. So far, there is no conclusive evidence suggesting that the moon had active volcanic activity after 3 billion years ago., \\
3. Observations and explorations of the lunar surface, including the density and distribution of impact craters, support the view that lunar volcanic activity ceased billions of years ago., \\
4. Compared to Earth, the moon's geological activity is very weak, and its rapid internal heat loss led to an earlier end to volcanic activity., \\
5. Research by NASA and other international space agencies supports the presence of ancient lunar volcanic activity, which had essentially ended by 3 billion years ago.], \\
\textbf{label\_p}: 0, \\
\textbf{Error Type}: \textcolor{red}{E:Factual Inconsistency} \\

\end{promptbox}
\clearpage

\begin{promptbox}[Chinese Example of Factuality Hallucination]{softblue}

\textbf{id}: 199, \\
\textbf{claim}: 吉林省发生了一起刑事案件，一名屠夫因为土地纠纷杀害了12人。, \\
\textbf{publish\_date}: 2024/8/14, \\
\textbf{label}: 1, \\
\textbf{source}: 中国互联网联合辟谣平台, \\
\textbf{gold evidence}: 经警方调查发现，案发地并未发生此类案件，这则消息纯属子虚乌有，系网民为了博取关注、吸引流量而编造发布的谣言。, \\
\textbf{domain}: 社会, \\
\textbf{model}: Baichuan4-Turbo, \\
\textbf{analysis}: [ \\
1. 吉林省确实发生了这起刑事案件,该事件在当地引起了广泛关注。, \\
2. 据警方通报，犯罪嫌疑人因土地纠纷产生不满情绪，最终导致了悲剧的发生。, \\
3. 被害人数为12人，这一数字得到了官方确认。, \\
4. 犯罪嫌疑人为一名屠夫，这一点也得到了官方证实。, \\
5. 此案件正在进一步调查中，相关责任人将依法受到严惩。], \\
\textbf{label\_p}: 0, \\
\textbf{Error Type}: \textcolor{red}{D:Factual Fabrication} \\

\textbf{id}: 1394, \\
\textbf{claim}: 月球的火山活动在 30 亿年前就停止了。, \\
\textbf{publish\_date}: 2023/1/20, \\
\textbf{label}: 1, \\
\textbf{source}: 科普中国, \\
\textbf{gold evidence}: 发现月球的火山活动其实一直持续到了 20.3亿年前，并不是之前认为的 30 亿年前就死了。, \\
\textbf{domain}: 科学, \\
\textbf{model}: Baichuan4-Turbo, \\
\textbf{analysis}: [ \\
1. 科学家们通过分析阿波罗任务带回的月球岩石样本发现，月球上的火山活动确实在约30亿年前达到了高峰，之后逐渐减少。, \\
2. 迄今为止，尚未有确凿证据表明月球在30亿年后仍有活跃的火山活动。, \\
3. 通过对月球表面的观测和探测，包括撞击坑的密度和分布等特征，支持月球火山活动在数十亿年前就已经停止的观点。, \\
4. 月球的地质活动与地球相比非常微弱，其内部热量散失快，导致火山活动较早结束。, \\
5. 美国宇航局（NASA）和其他国际空间机构的研究均支持月球在远古时期存在火山活动，而这些活动在30亿年前已经基本结束。], \\
\textbf{label\_p}: 0, \\
\textbf{Error Type}: \textcolor{red}{E:Factual Inconsistency} \\
\end{promptbox}

\clearpage

\begin{promptbox}[English Example of Reasoning Inadequacy]{softblue}

\textbf{id}: 2128, \\
\textbf{claim}: Smelling sachets or essential oils can help improve reduced sense of smell after COVID-19 recovery., \\
\textbf{publish\_date}: 2023/1/18, \\
\textbf{label}: 0, \\
\textbf{source}: CCTV News Client, \\
\textbf{gold evidence}: Doctor suggested using strong-smelling items to stimulate the sense of smell., \\
\textbf{domain}: Public Health, \\
\textbf{model}: GPT-3.5-Turbo, \\
\textbf{analysis}: [ \\
1. Smelling sachets or essential oils lacks scientific evidence as a treatment for reduced sense of smell., \\
2. Reduced sense of smell may indicate underlying pathological issues and requires professional diagnosis., \\
3. Home remedies like sachets or essential oils might delay necessary medical treatment., \\
4. Professional medical evaluation is critical for addressing symptoms effectively., \\
5. Unverified remedies should be approached with caution to avoid potential health risks.], \\
\textbf{label\_p}: 1, \\
\textbf{Error Type}: \textcolor{red}{G:Under Informativeness} \\

\textbf{id}: 28, \\
\textbf{claim}: Wuxi police recently arrested human traffickers., \\
\textbf{publish\_date}: 2024/10/15, \\
\textbf{label}: 1, \\
\textbf{source}: Chongqing Rumor Refutation, \\
\textbf{gold evidence}: Verified by police, no recent cases of child trafficking have occurred locally., \\
\textbf{domain}: Society, \\
\textbf{model}: GPT-4o, \\
\textbf{analysis}: [ \\
1. Wuxi police have a history of combating human trafficking and solving related cases., \\
2. Combating human trafficking is a key priority for China's security agencies., \\
3. China's Ministry of Public Security organizes nationwide operations against human trafficking., \\
4. Media and police frequently report on human trafficking arrests, including in the Wuxi region., \\
5. Human trafficking is a global problem, and China has implemented effective measures to address it.], \\
\textbf{label\_p}: 0, \\
\textbf{Error Type}: \textcolor{red}{F:Overgeneralized Reasoning} \\

\end{promptbox}

\clearpage

\begin{promptbox}[Chinese Example of Reasoning Inadequacy]{softblue}

\textbf{id}: 2128, \\
\textbf{claim}: 新冠康复后嗅觉减退可以通过闻香囊、精油等锻炼。, \\
\textbf{publish\_date}: 2023/1/18, \\
\textbf{label}: 0, \\
\textbf{source}: 央视新闻客户端, \\
\textbf{gold evidence}: 张伯礼：嗅觉、味觉不好的时候，可以闻香囊、鲜花，还有精油、风油精等。可以闻一闻刺激性强的气味锻炼嗅觉。, \\
\textbf{domain}: 公卫, \\
\textbf{model}: GPT-3.5-Turbo, \\
\textbf{analysis}: [ \\
1. 闻香囊、精油等并不能缓解嗅觉减退的症状。, \\
2. 嗅觉减退是一种病理状态，通常由多种原因引起，如感冒、鼻窦炎、脑部疾病等，需要及时就医进行诊断和治疗。, \\
3. 闻香囊、精油等虽然在一定程度上可以帮助人们放松情绪，但并没有科学证据表明它们可以治疗或改善嗅觉减退症状。, \\
4. 嗅觉减退通常需要通过专业的医学检查和治疗来解决，自行使用闻香囊、精油等方法不仅无效，还可能延误病情。, \\
5. 在出现嗅觉减退等健康问题时，应及时就医，接受专业医生的诊断和治疗，避免轻信未经科学验证的偏方或谣言。], \\
\textbf{label\_p}: 1, \\
\textbf{Error Type}: \textcolor{red}{G:Under Informativeness} \\

\textbf{id}: 28, \\
\textbf{claim}: 巫溪警方最近抓捕了人贩子。, \\
\textbf{publish\_date}: 2024/10/15, \\
\textbf{label}: 1, \\
\textbf{source}: 重庆辟谣, \\
\textbf{gold evidence}: 经巫溪警方核实，当地近期并未发生贩卖儿童案件，也未组织抓捕人贩子行动，网传“抓人贩子”系谣言。, \\
\textbf{domain}: 社会, \\
\textbf{model}: GPT-4o, \\
\textbf{analysis}: [ \\
1. 巫溪警方在多次行动中积极打击人口贩卖犯罪，过去也曾多次成功破获相关案件。, \\
2. 中国公安机关一直将打击人口贩卖作为重要任务，并定期通报相关案件的抓捕情况。, \\
3. 根据中国公安部的公开信息，各地警方定期开展专项行动以打击人口拐卖犯罪。, \\
4. 媒体报道和警方通报中经常会有关于人口贩子的抓捕新闻，巫溪地区也不例外。, \\
5. 人口贩卖是全球性问题，中国政府和警方对此采取了积极措施，并取得了一定成效。], \\
\textbf{label\_p}: 0, \\
\textbf{Error Type}: \textcolor{red}{F:Overgeneralized Reasoning} \\  
\end{promptbox}

\end{CJK}
\end{document}